\documentclass[letterpaper, 10 pt, conference]{IEEEtran}
\usepackage{times}

\IEEEoverridecommandlockouts 

\usepackage[pdftex]{graphicx}
\graphicspath{{figures/}}
\DeclareGraphicsExtensions{.pdf,.jpeg,.png,.bmp}

\usepackage{graphics} 
\usepackage{epsfig} 
\usepackage{mathptmx} 
\usepackage{times} 
\usepackage{amsmath} 
\usepackage{esint}
\usepackage{amssymb}  
\usepackage{subfigure}
\usepackage{booktabs}
\usepackage{float}
\usepackage[percent]{overpic}
\usepackage{gensymb}

\usepackage{helvet}         
\usepackage{courier}        
\usepackage{type1cm}        
\usepackage{makeidx}         
\usepackage{graphicx}        
\usepackage{multicol}        
\usepackage[bottom]{footmisc}
\usepackage{multicol}
\usepackage{bm}
\usepackage{color} 
\usepackage[bookmarks=true]{hyperref} 
\usepackage[utf8]{inputenc}
\usepackage{algpseudocode}
\usepackage{algorithm}
\usepackage{subfigure}
\usepackage{tikz}
\newcommand{\dittotikz}{%
    \tikz{
        \draw [line width=0.12ex] (-0.2ex,0) -- +(0,0.8ex)
            (0.2ex,0) -- +(0,0.8ex);
        \draw [line width=0.08ex] (-0.6ex,0.4ex) -- +(-1.5em,0)
            (0.6ex,0.4ex) -- +(1.5em,0);
    }%
}

\newcommand*\rfrac[2]{{}^{#1}\!/_{#2}}

\definecolor{CommentPMN}{rgb}{0.0,0.7,0.0}
\definecolor{CommentMT}{rgb}{0,0.0,0.7}
\definecolor{CommentPP}{rgb}{1.0,0.0,0.0}
\definecolor{CommentLP}{rgb}{1.0,0.5,0.0}
\definecolor{Blue}{rgb}{0.0,0.0,0.7}
\newcommand{\commentthis}[3]{{{\color{#1} {\textbf{#2} \textbf{#3}}}}}
\newcommand{\ignore}[1]{}
\newcommand{\pmnnotes}[1] { \commentthis{CommentPMN}{Paul says: }{``#1"}}

\newcommand{\pp}{{\bf p}}

\newcommand{\doctitle}{DENSER Cities:\\A System for \underline{Dens}e \underline{E}fficient \underline{R}econstructions of Cities}
\title{\LARGE \bf \doctitle}

\author{Michael Tanner$^\dagger$ \and Pedro Pini\'{e}s$^\dagger$ \and Lina Mar\'{i}a Paz$^\dagger$ \and Paul Newman$^\dagger$
  \thanks{$^\dagger$Mobile Robotics Group}%
  \thanks{Department of Engineering Science}%
  \thanks{University of Oxford}%
  \thanks{17 Parks Road, Oxford}%
  \thanks{OX1 3PJ, United Kingdom}%
  \thanks{mtanner,ppinies,linapaz,pnewman@robots.ox.ac.uk}
}%

\begin{document}
\makeatletter
\g@addto@macro\@maketitle{
\setcounter{figure}{0}
\begin{figure}[H]
    \setlength{\linewidth}{\textwidth}
    \setlength{\hsize}{\textwidth}
    \centering
    \includegraphics[width=18cm]{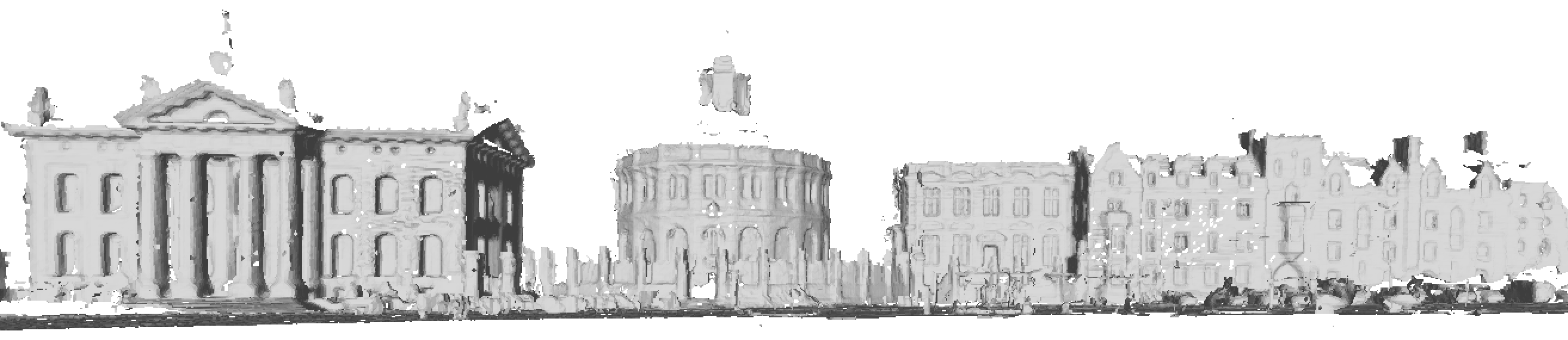}
    \caption{
    This paper is about the efficient generation of dense models of city-scale environments from range data.
    }
  \label{fig:money_maker}
    \end{figure}
}
\maketitle
\thispagestyle{empty}
\pagestyle{empty}

\begin{abstract}
This paper is about the efficient generation of dense, colored models of city-scale environments from range data and in particular, stereo cameras.
Better maps make for better understanding; better understanding leads to better robots, but this comes at a cost.
The computational and memory requirements of large dense models can be prohibitive. 

We provide the theory and the system needed to create city-scale dense reconstructions.
To do so, we apply a regularizer over a compressed 3D data structure while dealing with the complex boundary conditions this induces during the data-fusion stage.
We show that only with these considerations can we swiftly create neat, large, ``well behaved'' reconstructions.

We evaluate our system using the KITTI dataset and provide statistics for the metric errors in all surfaces created compared to those measured with 3D laser.
Our regularizer reduces the median error by 40\% in 3.4~km of dense reconstructions with a median accuracy of 6~cm.
For subjective analysis, we provide a qualitative review of 6.1~km of our dense reconstructions in an attached video.
These are the largest dense reconstructions from a single passive camera we are aware of in the literature.\\

Video:  \url{https://youtu.be/FRmF7mH86EQ}
\end{abstract}

\section{Introduction and Previous Work}

Over the past few years, the development of 3D reconstruction systems has undergone an explosion facilitated by the advances in GPU hardware.
Earlier, large-scale efforts such as \cite{pollefeys2008}\cite{FurukawaICCV2009}\cite{FurukawaCSS10} reconstructed sections of urban scenes from unstructured photo collections.
The ever-strengthening and broadening theoretical foundations of continuous optimization \cite{Chambolle:2011:FirstOrderPrimalDualAlgorithm}\cite{Goldluecke:2012:naturalvectorialtotal}, upon which the most advanced algorithms rely, have become accessible for robotics and computer vision applications.
Together these strands -- hardware and theory -- allow us to build systems which create city-scale 3D dense reconstructions.

However, the state of the art of many 3D reconstruction systems rarely considers scalability for the practical use in mapping applications such as autonomous driving or inspection.
The most general approaches are motivated by the recent mobile phone and tablet development \cite{Klingensmith_2015_7924}\cite{engel2015_stereo_lsdslam} with an eye on small-scale reconstruction.

We review the taxonomy of 3D reconstruction systems by considering the nature of the workspace to be modelled (e.g, indoors, outdoors) and the platform used for such a purpose.
We highlight the difference between dense reconstruction with a mobile robot \cite{vineet:etal:icra2015} versus object-centered modelling \cite{Newcombe:2011:KinectFusionRealtimedense}\cite{schoeps20153dv}.
The former suggests that the workspace is \emph{discovered} while being traversed --- e.g., when driving a car \cite{Whelan:2014:Realtimelargescaledense}\cite{vineet:etal:icra2015}\cite{engel2015_stereo_lsdslam}.
In contrast, in object-centred applications the goal is to generate models from sensor data gathered from carefully pre-selected viewpoints \cite{Curless:1996:volumetricmethodbuilding}\cite{Newcombe:2011:KinectFusionRealtimedense}\cite{Pradeep:2013:MonoFusionRealtime3D}.
The main difference is in the way the scene is observed and the interaction level of the sensor utilized with the environment --- object reconstruction requires a more \emph{active} interaction with the scene \cite{Pradeep:2013:MonoFusionRealtime3D}\cite{schoeps20153dv}.
In this paper we focus solely on \emph{passive} reconstruction. 

An important characteristic of 3D mapping algorithms is in their choice of representation. Some traditional Structure from Motion (SfM) and Simultaneous Localization and Mapping (SLAM) algorithms still believe in sparse and semi-dense representations with point clouds \cite{Alcantarilla14ppniv}\cite{mur2015orb}\cite{engel2015_stereo_lsdslam} or trust on probabilistic occupancy grids \cite{RSSYguel07} and mesh models \cite{pollefeys2008}.
Although these approaches have shown to be sufficient for accurate localisation, their maps lack the richness in representation that is prerequisite for a variety of applications, including robot navigation, active perception and semantic scene understanding.
In contrast, dense 3D reconstruction systems use the concept of the Truncated Signed Distance Function (TSDF) \cite{Curless:1996:volumetricmethodbuilding} to represent implicit surfaces from consecutive sensor observations.
To this end, we find data structures such as voxel grids \cite{Zeng:2013:Octreebasedfusionrealtime}, point based structures \cite{Keller:2013:Realtime3dreconstruction}, or more recently hashing voxel grids \cite{Teschner:2003:OptimizedSpatialHashing}\cite{vineet:etal:icra2015}\cite{schoeps20153dv}. 

The accuracy of the reconstruction also varies with the type of sensor utilized, ranging from monocular cameras, stereo cameras, RGB-D cameras \cite{Kim:2009:Multiviewimagetof}\cite{Newcombe:2011:DTAMDensetracking}\cite{Newcombe:2011:KinectFusionRealtimedense}\cite{Whelan:2014:Realtimelargescaledense}\cite{Zeng:2013:Octreebasedfusionrealtime}\cite{Klingensmith_2015_7924}\cite{vineet:etal:icra2015}\cite{schoeps20153dv}, to 3D laser \cite{RSSYguel07}.
Although very accurate models can be generated with RGB-D cameras, the quality of depth observations degrades with distance.
This fact restricts their use to small-scale and indoor work-spaces \cite{Klingensmith_2015_7924}\cite{Newcombe:2011:KinectFusionRealtimedense}\cite{Keller:2013:Realtime3dreconstruction} or marginally large indoor spaces \cite{Whelan:2014:Realtimelargescaledense}\cite{Zeng:2013:Octreebasedfusionrealtime}\cite{schoeps20153dv}.

Some techniques adapt themselves naturally to parallel architectures.
A popular approach followed by several stereo and monocular based algorithms splits the reconstruction problem up into two stages: initially, per-pixel depth maps are independently computed from images in a GPU. 
In a parallel thread, these depth maps are merged incrementally in a common reconstruction \cite{pollefeys2008}\cite{vineet:etal:icra2015}\cite{schoeps20153dv}.
Depth maps can be estimated by minimizing the photogrametric error over small image sequences or by estimating disparities from a triangulation on a set of support points \cite{Geiger2010ACCV}.
However, one must consider an important point: depth map estimates tend to be noisy and cannot deal well with sharp discontinuities. Therefore, these algorithms typically require additional post-processing steps with clever regularization to achieve better results \cite{Newcombe:2011:DTAMDensetracking}\cite{ranftl2012pushing}.
In this paper we push the limit on advanced convex optimization and regularization techniques for both depth-map estimation and fusion.

Our contributions are as follows.
First, we use a stereo model characterised by a second-order regularization energy term that has been successfully used in variational optical flow methods \cite{ferstl2014b}.
Our depth-map estimation approach is similar to \cite{ranftl2012pushing} where the regularization term is weighted and directed according to the input data by an anisotropic diffusion tensor, thus reducing the noise at object discontinuities.
Our choice allows sub-pixel accurate solutions and piecewise-planar depth maps.
To account for the effects of different illumination conditions, this model uses a robust similarity measure in the fidelity data term, the Ternary Census Signature (TCS) \cite{ZabihW94}.

Second, we present a method to regularize data stored in a compressed volumetric data structure, specifically the Hashing Voxel Grid \cite{Niessner:2013:Realtime3DReconstruction} --- thereby enabling optimal regularization of significantly larger scenes.
The key difficulty (and hence our vital contribution) in regularizing within a compressed structure is the presence of many additional boundary conditions introduced between regions which have and have not been directly observed by the range sensor.
Accurately computing the gradient and divergence operators, both of which are required to minimize the regularizer's energy, becomes a non-trivial problem.
Another subtle yet substantial problem is the way these conditions cause the regularizer term to inappropriately extrapolate surfaces, as noted in \cite{Tanner:2015:BOR2GBuildingOptimal}.

Finally, we provide, by way of our results, a new public-data benchmark for the community to compare dense-map reconstructions.
Our hope is that this becomes a tool of comparison within the dense mapping community.

\section{System Overview}\label{sec:system_overview}

\begin{figure}[t!]
  \centering{
    \includegraphics[width=1.0\columnwidth]{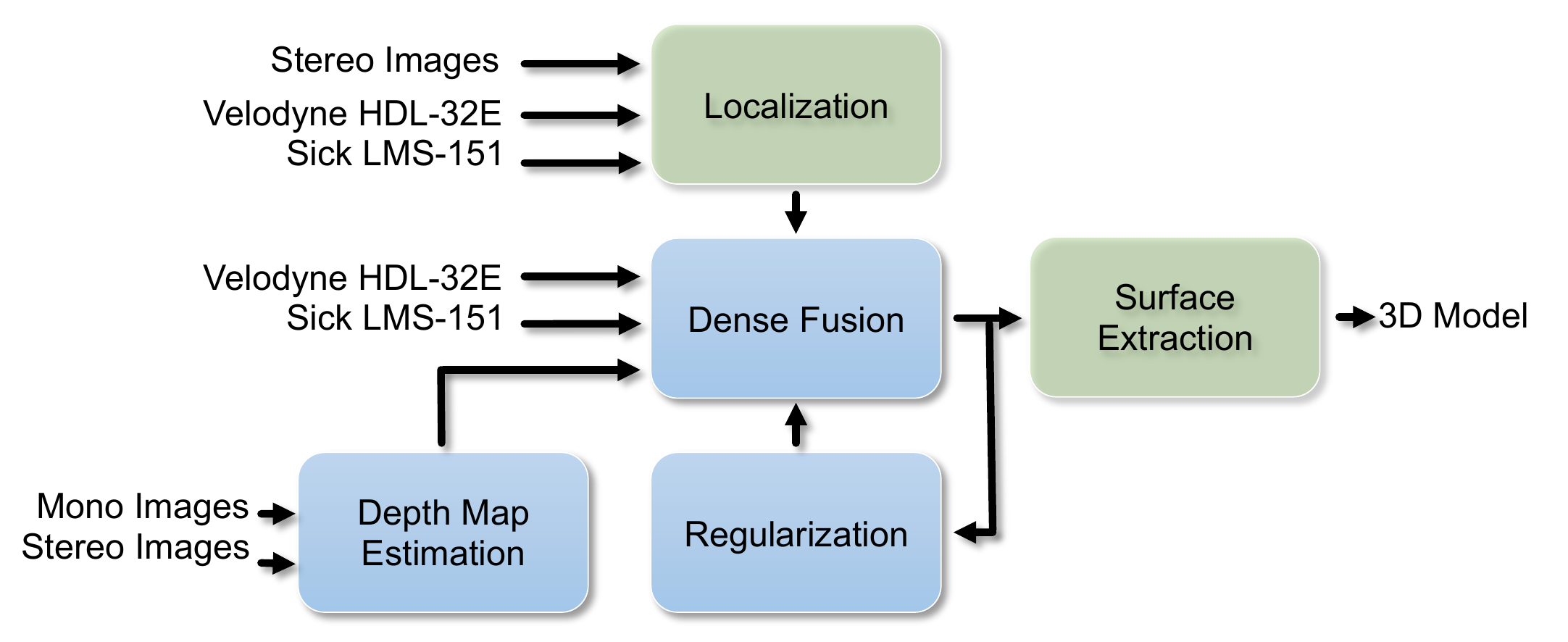}
  }
  \caption{
  An overview of our software pipeline.
  Our Dense Fusion module accepts range data from either laser sensors or depth maps created from mono or stereo cameras.
  We regularise the model in 3D space, extract the surface, and then provide the final 3D model to other components on our autonomous robotics platform --- e.g., segmentation, localisation, planning, or visualisation.
  The blue modules are discussed in further detail in Sections~\ref{sec:dense_mapper}-\ref{sec:depth_maps}.
  }
  \label{fig:pipeline}
\end{figure}

This section provides a brief overview of our system's architecture (Figure~\ref{fig:pipeline}) before we proceed to a detailed discussion in Sections~\ref{sec:dense_mapper} and \ref{sec:depth_maps}.

At its core, our system consumes range data and produces a 3D model.
For the passive reconstructions considered in this paper, this range information is primarily produced from stereo depth maps, nonetheless our system is implemented to augment the models with observations gathered from a variety of sensors --- specifically, we support calibrated stereo cameras, active cameras, and 2D/3D scanning lasers. 

We place no requirements on the trajectory of the source sensors; indeed, we illustrate our method using images captured from a forward-facing camera on a road vehicle --- an ill-conditioned use-case which is challenging yet likely given the utility of forward-facing cameras in navigation and perception tasks.
  
The pipeline consists of a localization module that is responsible for providing good estimates of the sensor pose where the observation is registered.
The design of this module is versatile allowing us to pick any arbitrarily SLAM derivative --- we use a standard approach that encompasses Stereo Visual Odometry along with loop-closure detection and a pose-graph optimization --- and as such, it is not the main focus of this paper.
The Depth Map Estimation module is responsible for processing stereo frames into a stream of denoised depth maps.
The Dense Fusion module merges the depth maps -- which are noisy indeed if they hail from a stereo pair -- into a compressed data structure that is continuously regularized.
New incoming data can be added at any time from any suitable sensor source.
Similarly a surface model can be extracted to be processed in parallel by a separate application (e.g., planner).

\section{3D Dense Maps}\label{sec:dense_mapper}

The core of the 3D dense mapping system consists of a Dense Fusion module which integrates a sequence of range observations into a volumetric representation, and the Regularisation module smooths noisy surfaces and removes uncertain surfaces.

\subsection{Fusing Data}

To create a dense reconstruction for a set of input range data, one must process the range values within a data structure which can efficiently represent surfaces and continually improve that representation with future observations.
Simply storing each of the range estimates (e.g., as a point cloud) is a poor choice as storage grows without bound and the surface reconstruction does not improve when \emph{revisiting} a location.

A common approach is to select a subset of space in which one will reconstruct surfaces and divide the space into a uniform voxel grid.
Each voxel stores range observations represented by their corresponding Truncated Signed Distance Function (TSDF), $u_{TSDF}$.
The voxels' TSDF values are computed such that one can solve for the zero-crossing level set (isosurface) to find a continuous surface model.
Even though the voxel grid is a discrete field, because the TSDF value is a real number, the surface reconstruction is even more precise than the voxel size.

Due to memory constraints, only a small subset of space can be reconstructed using a legacy approach where the grid is fixed in space and therefore reconstructs only a few cubic meters.
This presents a particular problem in mobile robotics applications since the robot's exploration region is restricted to a prohibitively small region.
In addition, long-range depth sensors (e.g., laser) cannot be fully utilised since their range exceeds the size of the voxel grid (or local voxel grid if a local-mapping approach is used) \cite{Whelan:2012:KintinuousSpatiallyExtended}.
A variety of techniques were proposed in recent years to remove these limits.
They leverage the fact the overwhelming majority of voxels do not contain any valid TSDF data since they were never directly observed by the range sensor.
A compressed data structure only allocates and stores data in voxels which are near a surface.
The most successful approach is the Hashing Voxel Grid (HVG)  \cite{Niessner:2013:Realtime3DReconstruction}.
The HVG subdivides the world into an infinite, uniform grid of voxel \emph{blocks}, each of which represents its own small voxel grid with 8 voxels in each dimension (512 total).
Anytime a surface is observed within a given voxel block, all the voxels in that block are allocated and their TSDF values are updated.
Blocks are only allocated where surfaces are observed.

\begin{figure}[t!]
  \centering{
    \includegraphics[width=1.0\columnwidth]{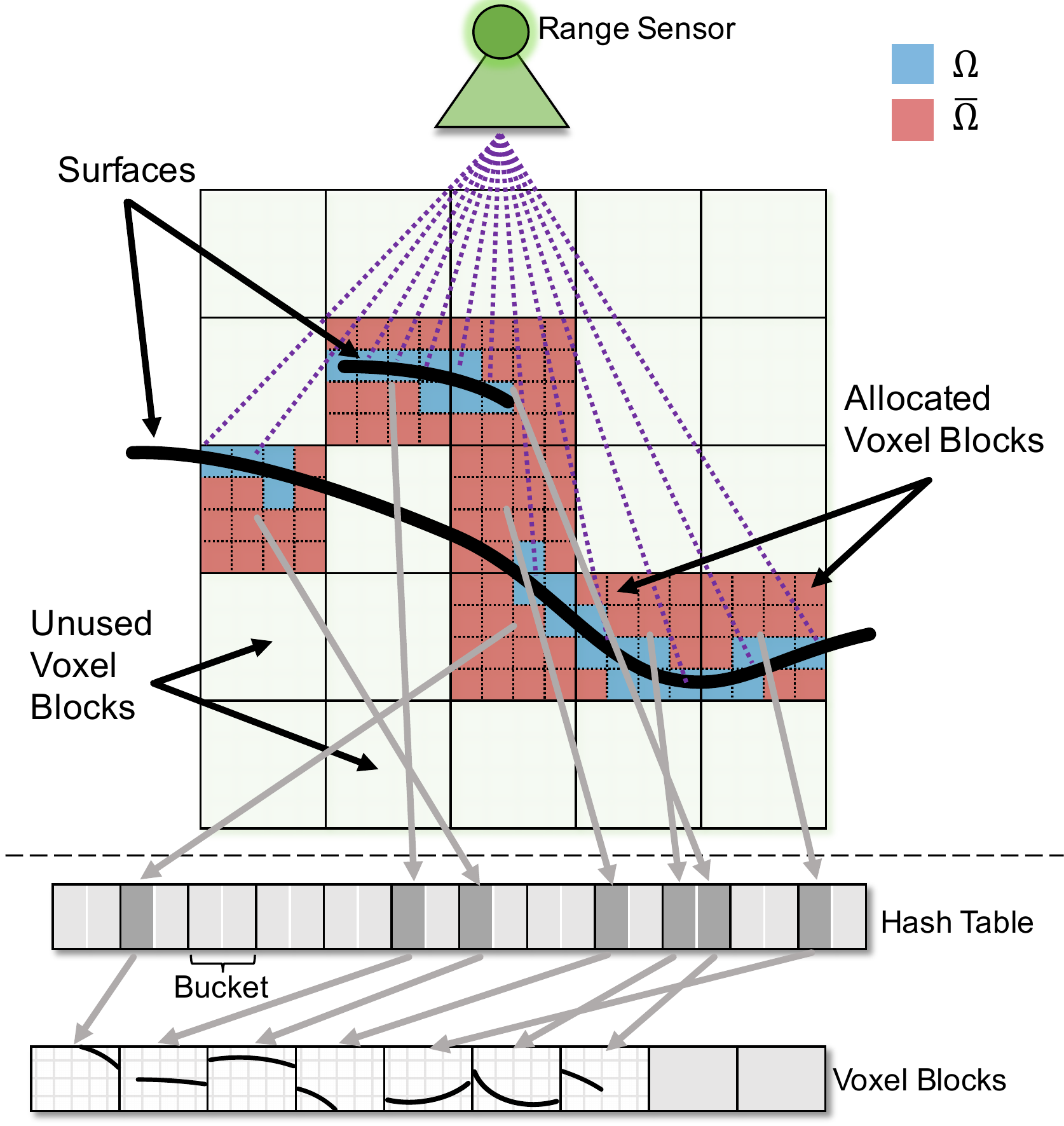}
  }
  \caption{
  A depiction of our novel combination of the Hashing Voxel Grid (HVG) data structure with regularization to fuse depth observations from the environment.
  The HVG enables us to reconstruct large-scale scenes by only allocating memory for the regions of space in which surfaces are observed (i.e., the colored blocks).
  To avoid generating spurious surfaces, we mark each voxel with an indicator variable ($\Omega$) to ensure the regularizer only operates on voxels in which depth information was directly observed.
  This same approach is used independent of the range sensor --- e.g., stereo depth maps or laser.
  Figure inspired by \cite{Niessner:2013:Realtime3DReconstruction}.
  }
  \label{fig:hvg_and_omega}
\end{figure}

Applying a hash function to coordinates in world space gives an index within the hash table, which in turn points to the raw voxel data.
Figure~\ref{fig:hvg_and_omega} provides a graphical overview of this process, and we refer the reader to the original HVG paper \cite{Niessner:2013:Realtime3DReconstruction} for further implementation details.

If one considers each HVG voxel block to be a legacy voxel grid, then the update equations are identical to those presented by \cite{Newcombe:2011:KinectFusionRealtimedense}.
For each voxel, perform the following operations for every new depth map, $D$:

\begin{enumerate}
  \item Calculate the voxel's global-frame center $\pp_g =[x_g,y_g,z_g]^T$ with respect to the camera coordinate frame as $\pp_c = \mathbf{T}_{gc}^{-1} \pp_g$

  \item Project $\pp_c$ into $D$ to determine the nearest pixel $d_{x,y}$.

  \item If the pixel $(x,y)$ lies within the depth map, evaluate $u_{SDF} = d_{x,y} - z_c$. $u > 0$ indicates the voxel is between the surface and the camera, and $u < 0$ for voxels occluded from the camera by the surface.

  \item Update the voxel's current $f$ (TSDF value) and $w$ (weight or confidence in the given TSDF value),

    \begin{equation}
      \begin{split}
        w_k &=
        \begin{cases}
          w_{k-1} + 1  & u_{SDF} \geq -\mu \\
          w_{k-1}      & u_{SDF} < -\mu
        \end{cases}\\
        f_k &=
        \begin{cases}
          \frac{ u_{TSDF} + w_{k-1} f_{k-1} }{ w_{k} }  & u_{SDF} \geq -\mu \\
          f_{k-1}                                       & u_{SDF} < -\mu
        \end{cases}\\
      \end{split}
    \end{equation}

    where $w_{k-1}$ and $f_{k-1}$ are the previous values of $f$ and $w$ for that voxel.

\end{enumerate}

This  method projects voxels into the depth map to update $f$, but this can be extended to work with laser sensors by only updating voxels along the ray from the laser sensor to $-\mu$ behind the observed surface.

\begin{figure*}[t!]
  \centering{
    \includegraphics[width=1.8\columnwidth]{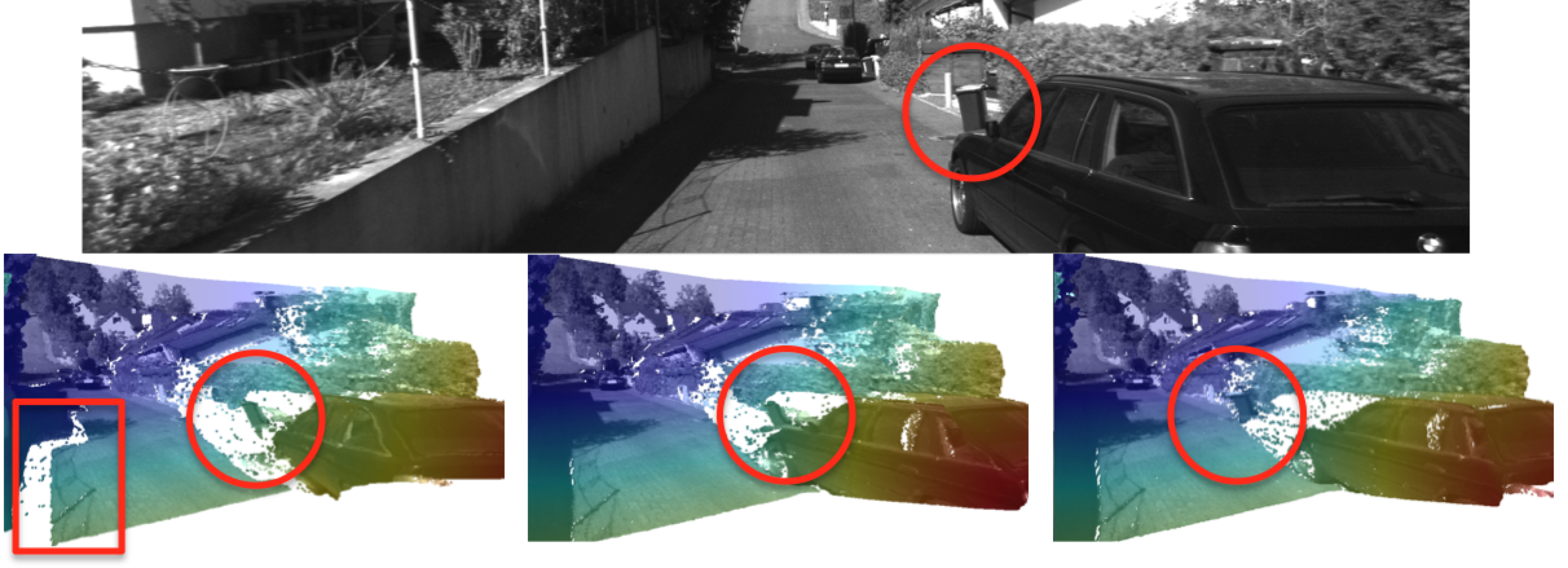}
  }
  \caption{
  Comparison of three depth-map regularizers: TV, TGV, and TGV-Tensor.
  Using the reference image (top), the TV regularizer (left) favours fronto-parallel surfaces, therefore it creates a sharp discontinuity for the shadow on the road (red rectangle) and attaches the rubbish bin to the rear of the car (red circle).
  TGV (center) improves upon this by allowing planes at any orientation, but it still cannot identify boundaries between objects:  the rubbish bin is again estimated as part of the car.
  Finally, the TGV-Tensor (right) regularizer both allows planes at any orientation and is more successful at differentiating objects by taking into account the normal of the color image's gradient. 
  For clarity, the reconstructions have a different viewing origin than the reference image.
  }
  \label{fig:tv_tgv_tensor}
\end{figure*}

\subsection{3D Regularization}

We pose the fusion step as a noise-reduction problem that can be approached by a continuous energy minimization over the voxel-grid domain ($\Omega$):

\begin{equation}\label{eq:vm_energy_tv_simple}
  \begin{split}
    E(u) &= E_{regularization}(u) + E_{data}(u,f) \\
    E(u) &= \int_\Omega | \nabla u | d\Omega +
    \lambda \int_\Omega ||f - u||_2^2 d\Omega
  \end{split}
\end{equation}

\noindent where $E(u)$ is the energy (which we seek to minimize) of the denoised ($u$) and noisy ($f$) TSDF data.
The $data$ energy term seeks to minimize the difference between the $u$ and $f$ while $\lambda$ controls the relative importance of the data term vs. the regularizer term. 
The $regularization$ energy term, commonly referred to as a Total Variation (TV) regularizer, seeks to fit the solution ($u$) to a specified prior --- the L1 norm in this case.
This is a convex energy minimization problem that can be solved using Primal-Dual techniques \cite{Chambolle:2011:FirstOrderPrimalDualAlgorithm}.

In practice, the TV norm has a two-fold effect: (1) smooths out the reconstructed surfaces, and (2) removes surfaces which are ``uncertain'' --- i.e., voxels with high gradients and few direct range observation.
However, since compressed voxel grid data structures are not regular in space, the proper method to compute the gradient (and its dual: divergence) in the presence of the additional boundary conditions is not straightforward --- hence why we believe this has remained an open problem.
In addition, improper gradient and divergence operators will cause the regulariser to spuriously extrapolate surfaces into undesired regions of the reconstruction. 
We leverage the legacy voxel grid technique \cite{Tanner:2015:BOR2GBuildingOptimal} into the HVG by introducing a new state variable in each voxel indicating whether or not it was directly observed by a range sensor.
The subset of voxels which were observed are defined as $\Omega$, the set solely upon which the regluarizer is constrained to operate, thus avoiding spurious surface generation in non-valid voxels ($\bar\Omega$).
Note that all voxel \emph{blocks} which are \emph{not} allocated are in $\bar\Omega$.
A graphical depiction of the $\Omega$ and $\bar\Omega$ for a sample surface is provided in Figure~\ref{fig:hvg_and_omega}.

The additional boundary conditions introduced by the HVG require careful derivation of the gradient (Equation~\ref{eq:modified_gradient}) and divergence (Equation~\ref{eq:modified_divergence}) operators to take into account whether or not neighbors are in $\Omega$.
We define the gradient as:

\begin{equation}\label{eq:modified_gradient}
\nabla_x{u_{i,j,k}} = \begin{cases}
        u_{i+1,j,k} - u_{i,j,k} & \text{if~} 1 \leq i < V_x\\
        0                       & \text{if~} i = V_x\\
        0                       & \text{if~} u_{i,j,k} \in \bar\Omega\\
        0                       & \text{if~} u_{i+1,j,k} \in \bar\Omega\\
        \end{cases}
\end{equation}

\noindent where $u_{i,j,k}$ is a voxel's TSDF value at the 3D world integer coordinates $(i,j,k)$, and $V_x$ is the number of voxels in the $x$ dimension.
The gradient term in the regularizer encourages smoothness across neighbouring voxels which explains why this new gradient definition excludes $\bar\Omega$ voxels  --- they have not been observed.

To solve the primal's dual optimization problem (Section~\ref{sec:3d_reg_implementation}), we define the divergence operator as the dual of the new gradient operator:

\begin{equation}\label{eq:modified_divergence}
\nabla_x\cdot{\pp_{i,j,k}} = \begin{cases}
        \pp^x_{i,j,k} - \pp^x_{i-1,j,k}  & \text{if~} 1 < i < V_x\\
        \pp^x_{i,j,k}                        & \text{if~} i = 1\\
        -\pp^x_{i-1,j,k}                     & \text{if~} i = V_x\\
        0                                    & \text{if~} u_{i,j,k} \in \bar\Omega\\
        \pp^x_{i,j,k   }                     & \text{if~} u_{i-1,j,k} \in \bar\Omega\\
        -\pp^x_{i-1,j,k}                     & \text{if~} u_{i+1,j,k} \in \bar\Omega\\
        \end{cases}
\end{equation}

Each voxel block is treated as a voxel grid with boundary conditions which are determined by its neighbours' $\Omega$ indicator function, $I_\Omega$.
The regularizer operates only on the voxels within $\Omega$, the domain of integration, and thus it neither spreads spurious surfaces into unobserved regions nor updates valid voxels with invalid TSDF data.
Note that both equations are presented for the $x$-dimension, but the $y$ and $z$-dimension equations can be obtained by variable substitution between $i$, $j$, and $k$.

\subsection{Implementation of 3D Energy Minimization}\label{sec:3d_reg_implementation}

In this section, we describe the algorithm to solve Equation~\ref{eq:vm_energy_tv_simple} and point the reader to \ignore{\cite{Pinies:2015:TooMuchTV}}\cite{Chambolle:2011:FirstOrderPrimalDualAlgorithm}\cite{Rockafellar:1970:ConvexAnalysis} for a detailed derivation of these steps.
We vary from their methods only in our new definition for the gradient and divergence operators.

Equation~\ref{eq:vm_energy_tv_simple} is not smooth so it cannot be minimized with traditional techniques.
We convert the TV term to a differentiable form via the Legendre-Fenchel Transform \cite{Rockafellar:1970:ConvexAnalysis} and then use the Proximal Gradient method \cite{Chambolle:2011:FirstOrderPrimalDualAlgorithm} to transform our TV cost-function term into:

\begin{equation}
  \min_u \int_\Omega |\nabla{u}| d\Omega =
  \min_u \max_{||\pp||_\infty\leq 1} \int_\Omega u\nabla \cdot \pp d\Omega
\end{equation}

\noindent where the primal scalar $u$ is the current denoised and interpolated TSDF solution and $\nabla \cdot \pp$ is the divergence of the dual vector field, $\nabla\cdot\pp = \nabla{p}_x + \nabla{p}_y + \nabla{p}_z$. Substituted into Equation~\ref{eq:vm_energy_tv_simple}, it becomes a saddle-point problem to maximise the new dual variable $\pp$ while minimizing the original primal variable $u$,

\begin{equation}
  \min_u \max_{||\pp||_\infty\leq 1} \int_\Omega u \nabla \cdot \pp + \lambda \int_\Omega ||f-u||_2^2 d\Omega
\end{equation}

This can be efficiently solved via a Primal-Dual optimization algorithm \cite{Chambolle:2011:FirstOrderPrimalDualAlgorithm}:

\begin{enumerate}
  \item $\pp$, $u$, and $\hat{u}$ are initialised to $0$.  $\hat{u}$ is a temporary variable which reduces the number of optimization iterations required to converge.

  \item To solve the maximisation, update the dual variable $\pp$,

    \begin{equation}
      \begin{split}
        \pp_k  &= \frac{ \tilde{\pp} }{ \max(1,||\tilde{\pp}||_2) } \\
        \tilde{\pp}  &= \pp_{k-1} + \sigma_p \nabla \hat{u}
      \end{split}
    \end{equation}

    where $\sigma_p$ is the dual variable's gradient-ascent step size.

  \item Then update $u$ to minimize the primal variable,

    \begin{equation}
    \begin{split}
      u_k       &= \frac{ \tilde{u} + \tau \lambda w f }{ 1 + \tau \lambda w } \\
      \tilde{u} &= u_{k-1} - \tau\nabla\cdot\pp
    \end{split}
    \end{equation}
    
    where $\tau$ is the gradient-descent step size and $w$ is the weight of the $f$ TSDF value.
    
  \item Finally, the energy converge in fewer iterations with a ``relaxation'' step,

    \begin{equation}\label{eq:relax}
      \hat{u} = u + \theta(u - \hat{u})
    \end{equation}

    where $\theta$ is a parameter to adjust the relaxation step size.

\end{enumerate}

As the operations in each voxel are independent, our implementation leverages parallel GPU computing with careful synchronization between subsequent primal and dual variable updates.

\begin{table*}[t!]
\centering
\caption{Summary of Parameters Used in System}
\label{tab:parameters}
\begin{tabular}{p{1cm}p{4cm}p{11cm}}
\toprule
\textbf{Symbol}          & \textbf{Value}                & \textbf{Description} \\
\midrule
$\lambda_{3D}$           & 0.8 (10~cm) / 0.4 (20~cm)     & The Total Variation (TGV) weighting of the data term vs. regularization \\
$\mu_{3D}$               & 1.0~m (10~cm) / 1.6~m (20~cm) & The maximum voxel distance behind a surface in which to fuse negative signed-distance values \\
$\sigma_p$, $\theta$     & 0.5, 1.0                      & 3D regularizer gradient-ascent/descent step sizes \\
$\tau$                   & $\rfrac{1}{6}$                & 3D regularizer relaxation-step weight \\
\midrule
$\lambda_{2D}$           & 0.5                           & The Total Global Variation (TGV) weighting of the data term vs. regularization \\
$\alpha_{1}, \alpha_{2}$ & 1.0, 5.0                      & Relative weights of the affine-smooth/piecewise-constant TGV terms \\
$\beta$, $\gamma$        & 1.0, 4.0                      & Exponent and scale factor respectively modifying the influence of the image gradient\\ \bottomrule
\end{tabular}
\end{table*}

\begin{table}[t!]
\centering
\caption{Summary of KITTI-VO Scenarios and Fusion Time For 10 \& 20~cm Voxels}
\label{tab:kitti_vo_summary}
\begin{tabular}{@{}rrrrrrr@{}}
\toprule
\textbf{KITTI-VO \#} & \textbf{Length (km)} & \textbf{\# Frames} & \textbf{Fusion Time (mm:ss)} \\ \midrule
Sequence 00          & 1.0                  & 1419               & 4:10 / 9:21 \\
Sequence 07          & 0.7                  & 1101               & 3:23 / 7:13 \\
Sequence 09          & 1.7                  & 1591               & 4:09 / 9:18 \\ \bottomrule
\end{tabular}
\end{table}

\begin{table*}[t!]
\centering
\caption{Summary of Error Statistics (10~cm voxels)}
\label{tab:kitti_vo_errors_10cm}
\begin{tabular}{@{}rrrrrrrrrr@{}}
\toprule
\textbf{KITTI-VO \#} & \textbf{Type} & \textbf{Mode (cm)} & \textbf{Median (cm)} & \textbf{75\% (cm)} & \textbf{GPU Memory} & \textbf{Surface Area} & \textbf{\# Voxels 10$^6$} & \textbf{Time Per Iter. (mm:ss)}\\ \midrule
Sequence 00          & Raw           & 0.84               & 10.00                & 36.51              & 976 MiB             & 51010~m$^2$           & 123.62                    & \\
\dittotikz           & Regularized   & 1.84               & 6.15                 & 23.60              & \dittotikz          & 34630~m$^2$           & \dittotikz                & 5:24 \\
Sequence 07          & Raw           & 1.68               & 12.19                & 40.38              & 637 MiB             & 33891~m$^2$           & 79.11                     & \\
\dittotikz           & Regularized   & 1.69               & 7.30                 & 26.21              & \dittotikz          & 22817~m$^2$           & \dittotikz                & 4:10 \\
Sequence 09          & Raw           & 2.00               & 8.43                 & 32.22              & 1,462 MiB           & 81624~m$^2$           & 187.24                    & \\
\dittotikz           & Regularized   & 1.83               & 5.04                 & 19.06              & \dittotikz          & 54561~m$^2$           & \dittotikz                & 4:41 \\
\bottomrule
\end{tabular}
\end{table*}

\begin{table*}[t!]
\centering
\caption{Summary of Error Statistics (20~cm voxels)}
\label{tab:kitti_vo_errors_20cm}
\begin{tabular}{@{}rrrrrrrrr@{}}
\toprule
\textbf{KITTI-VO \#} & \textbf{Type} & \textbf{Mode (cm)} & \textbf{Median (cm)} & \textbf{75\% (cm)} & \textbf{GPU Memory} & \textbf{Surface Area} & \textbf{\# Voxels (10$^6$)} & \textbf{Time Per Iter. (mm:ss)}\\ \midrule
Sequence 00          & Raw           & 1.84               & 10.47                & 39.04              & 221 MiB             & 45830~m$^2$           & 24.66                       & \\
\dittotikz           & Regularized   & 1.85               & 6.20                 & 23.02              & \dittotikz          & 30696~m$^2$           & \dittotikz                  & 0:19 \\
Sequence 07          & Raw           & 1.68               & 12.43                & 42.50              & 157 MiB             & 30585~m$^2$           & 16.23                       & \\
\dittotikz           & Regularized   & 1.84               & 7.22                 & 25.13              & \dittotikz          & 20394~m$^2$           & \dittotikz                  & 0:22 \\
Sequence 09          & Raw           & 2.00               & 8.45                 & 33.06              & 332 MiB             & 74015~m$^2$           & 39.13                       & \\
\dittotikz           & Regularized   & 1.84               & 5.13                 & 18.38              & \dittotikz          & 48511~m$^2$           & \dittotikz                  & 1:04 \\
\bottomrule
\end{tabular}
\end{table*}

\begin{figure*}[t!]
  \centering{
    \subfigure[Sequence 00 (Without regularization)]{
    \includegraphics[width=0.45\columnwidth]{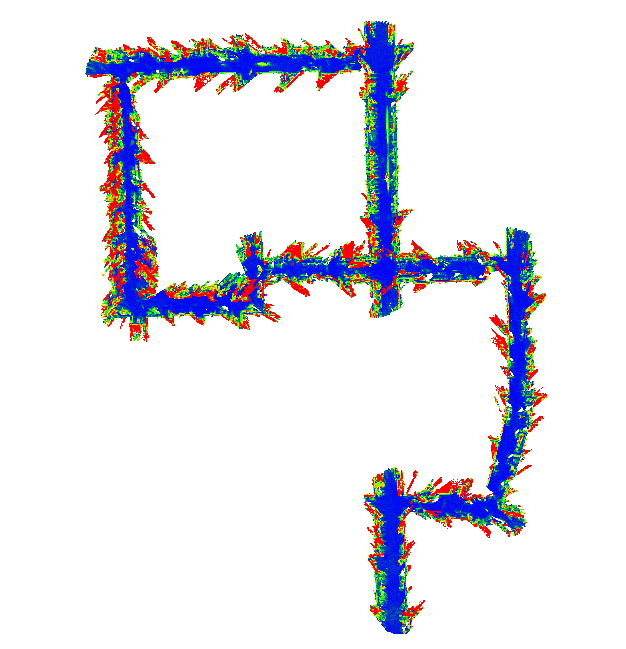}
    \begin{overpic}[width=0.5\columnwidth]{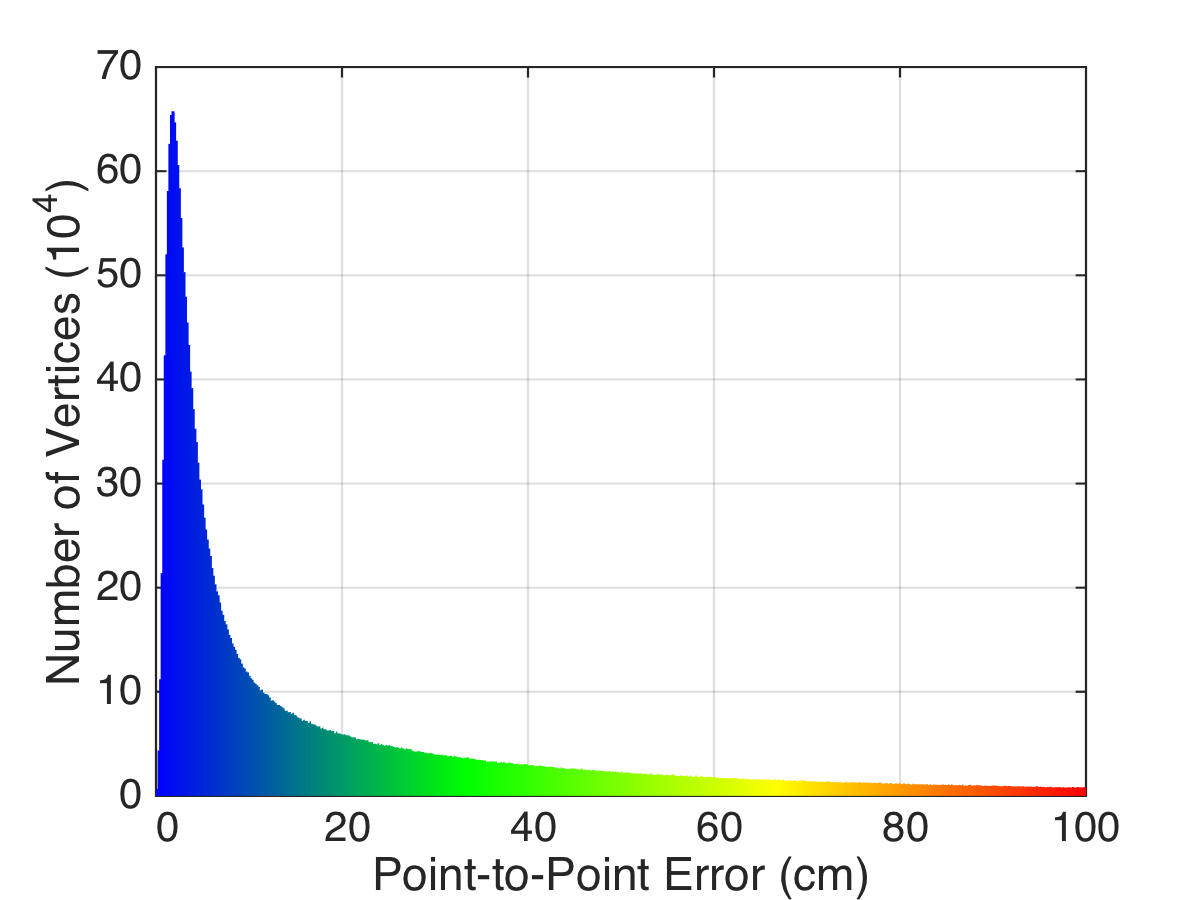}
      \put(30,60){\footnotesize median $=$ 10.00~cm}
      \put(30,50){\footnotesize 75\% $=$ 36.51~cm}
    \end{overpic}
    }
    \subfigure[Sequence 00 (With regularization)]{
    \begin{overpic}[width=0.5\columnwidth]{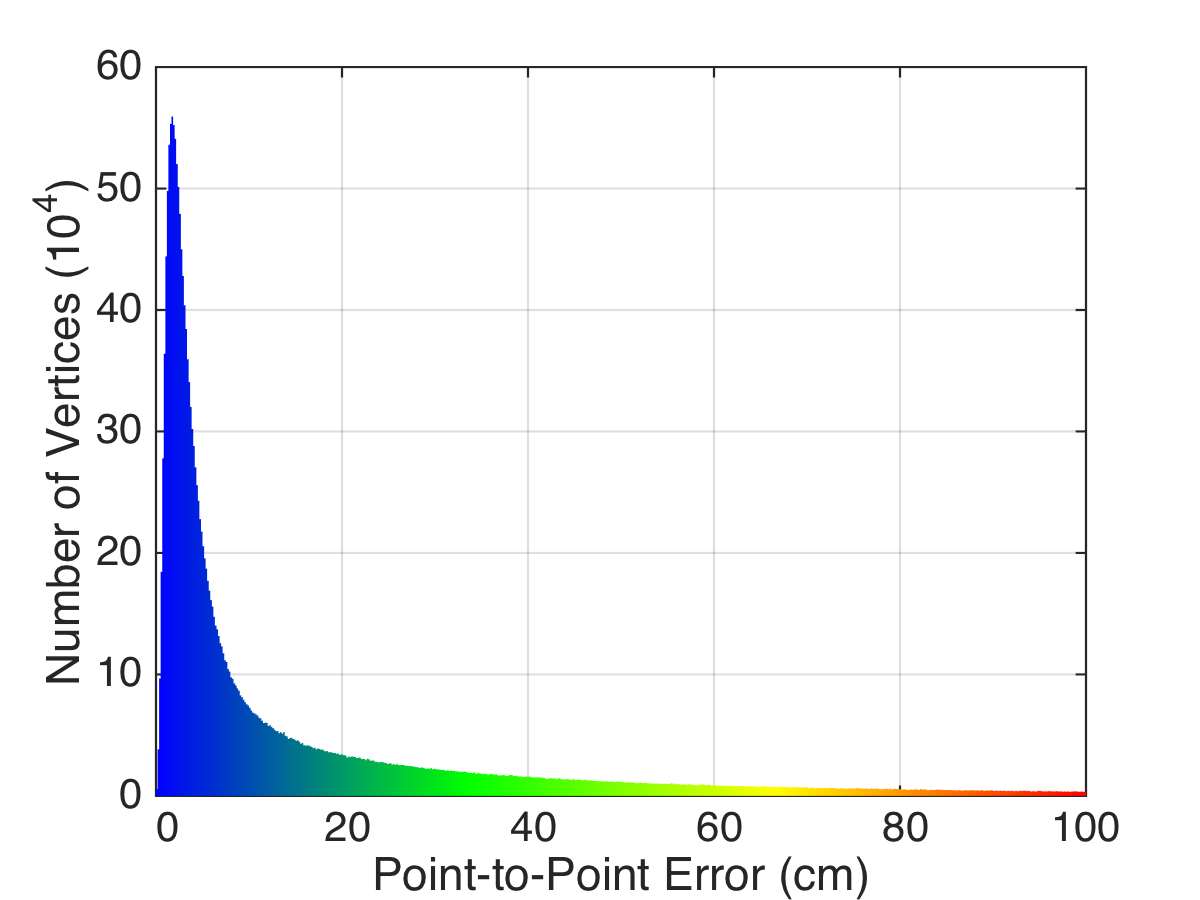}
      \put(30,60){\footnotesize median $=$ 6.15~cm}
      \put(30,50){\footnotesize 75\% $=$ 23.60~cm}
    \end{overpic}
    \includegraphics[width=0.45\columnwidth]{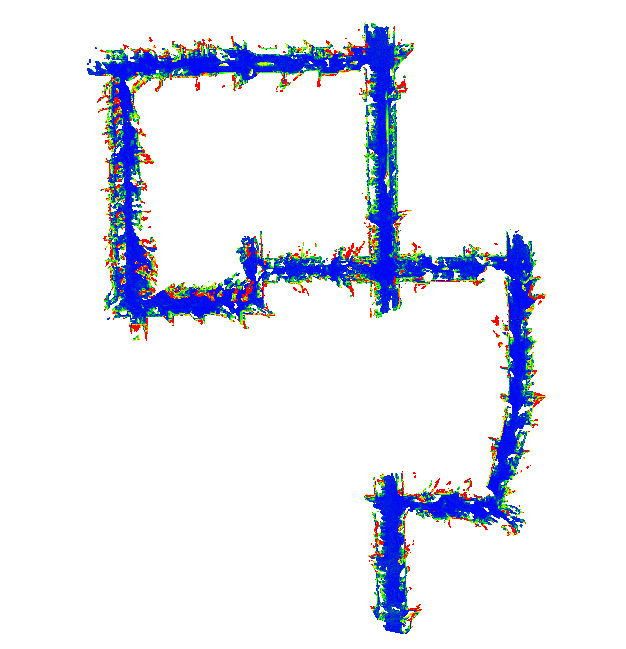}
    }
    
    \subfigure[Sequence 07 (Without regularization)]{
    \includegraphics[width=0.45\columnwidth]{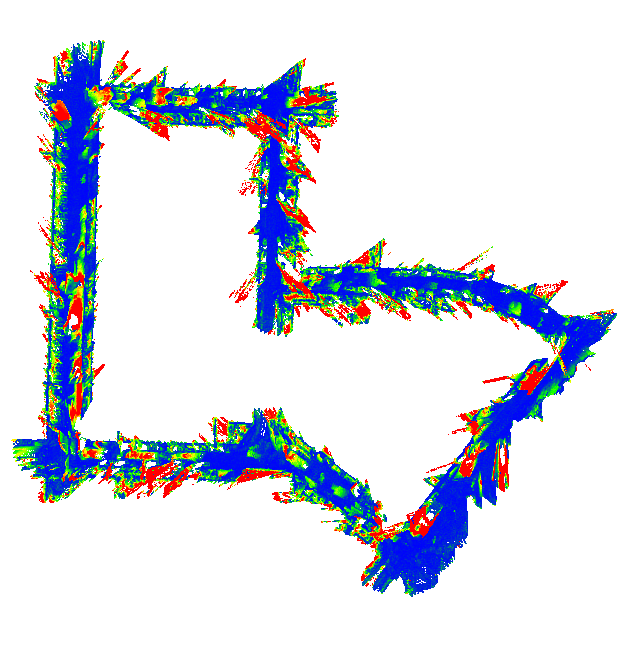}
    \begin{overpic}[width=0.5\columnwidth]{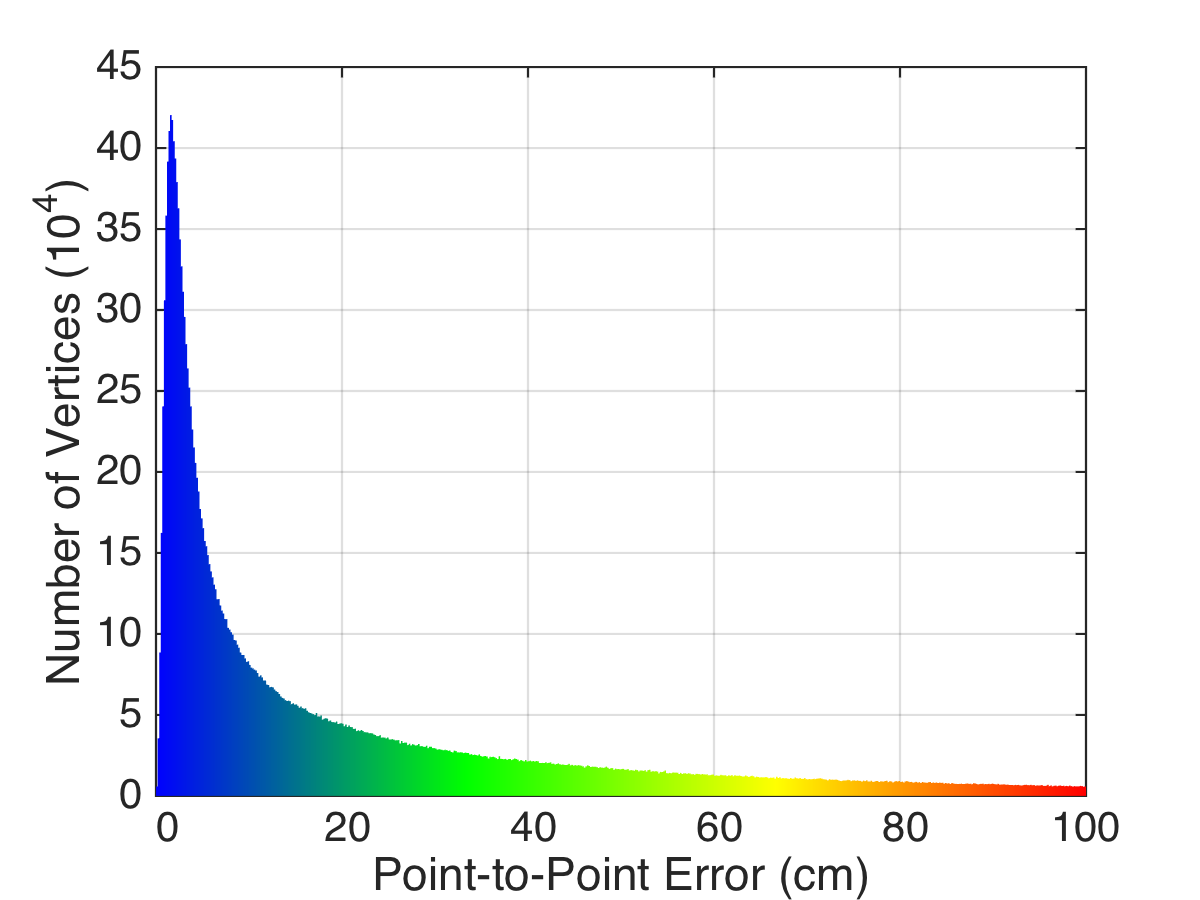}
      \put(30,60){\footnotesize median $=$ 12.19~cm}
      \put(30,50){\footnotesize 75\% $=$ 40.38~cm}
    \end{overpic}
    }
    \subfigure[Sequence 07 (With regularization)]{
    \begin{overpic}[width=0.5\columnwidth]{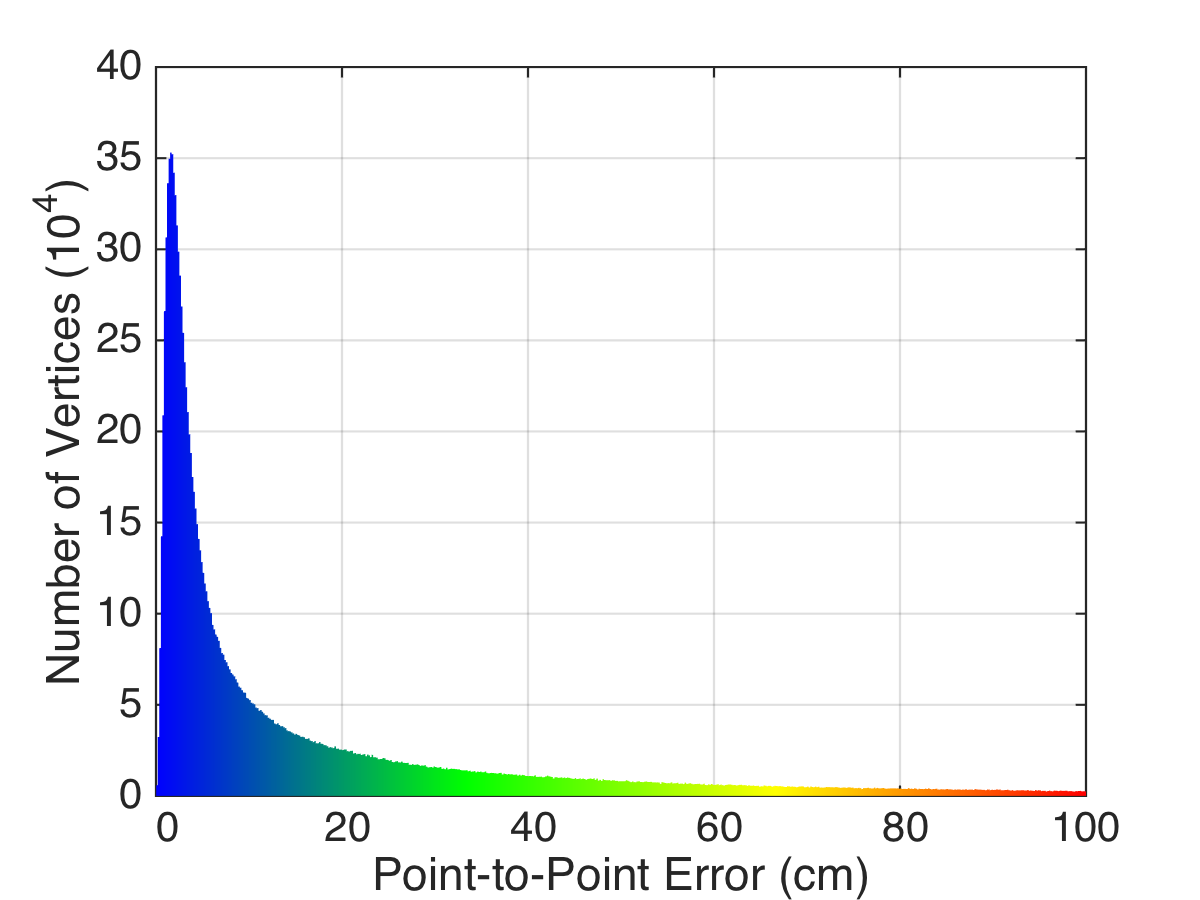}
      \put(30,60){\footnotesize median $=$ 7.30~cm}
      \put(30,50){\footnotesize 75\% $=$ 26.21~cm}
    \end{overpic}
    \includegraphics[width=0.45\columnwidth]{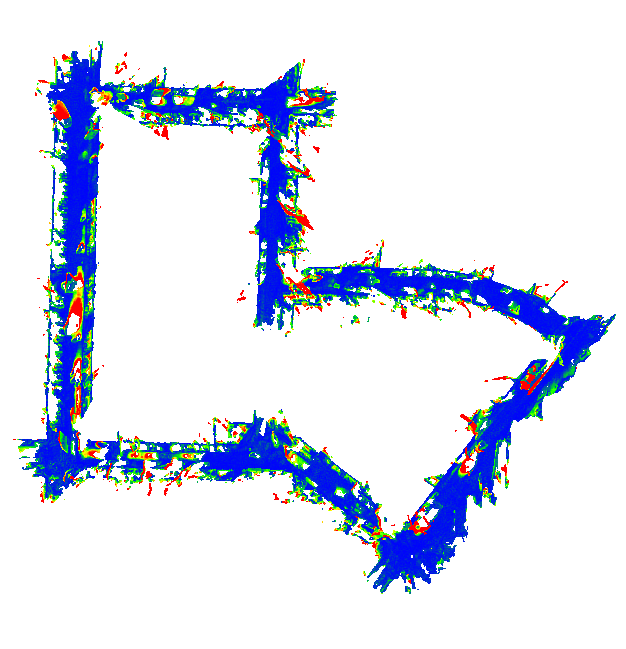}
    }
    
    \subfigure[Sequence 09 (Without regularization)]{
    \includegraphics[width=0.45\columnwidth]{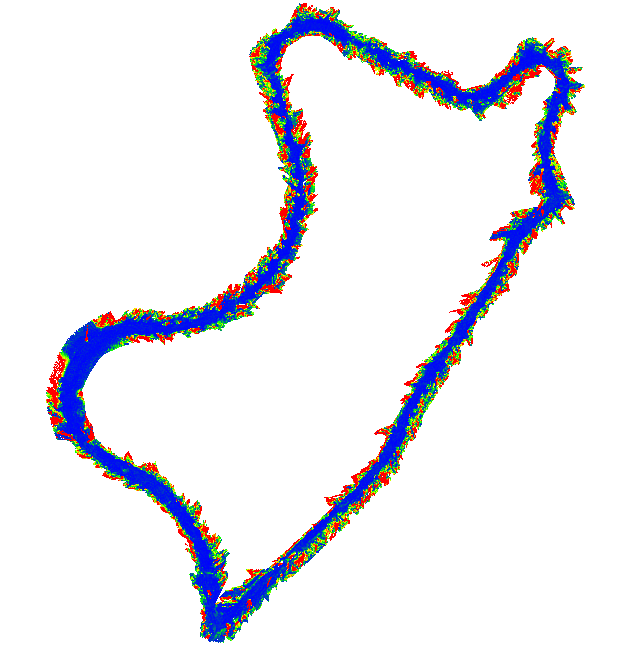}
    \begin{overpic}[width=0.5\columnwidth]{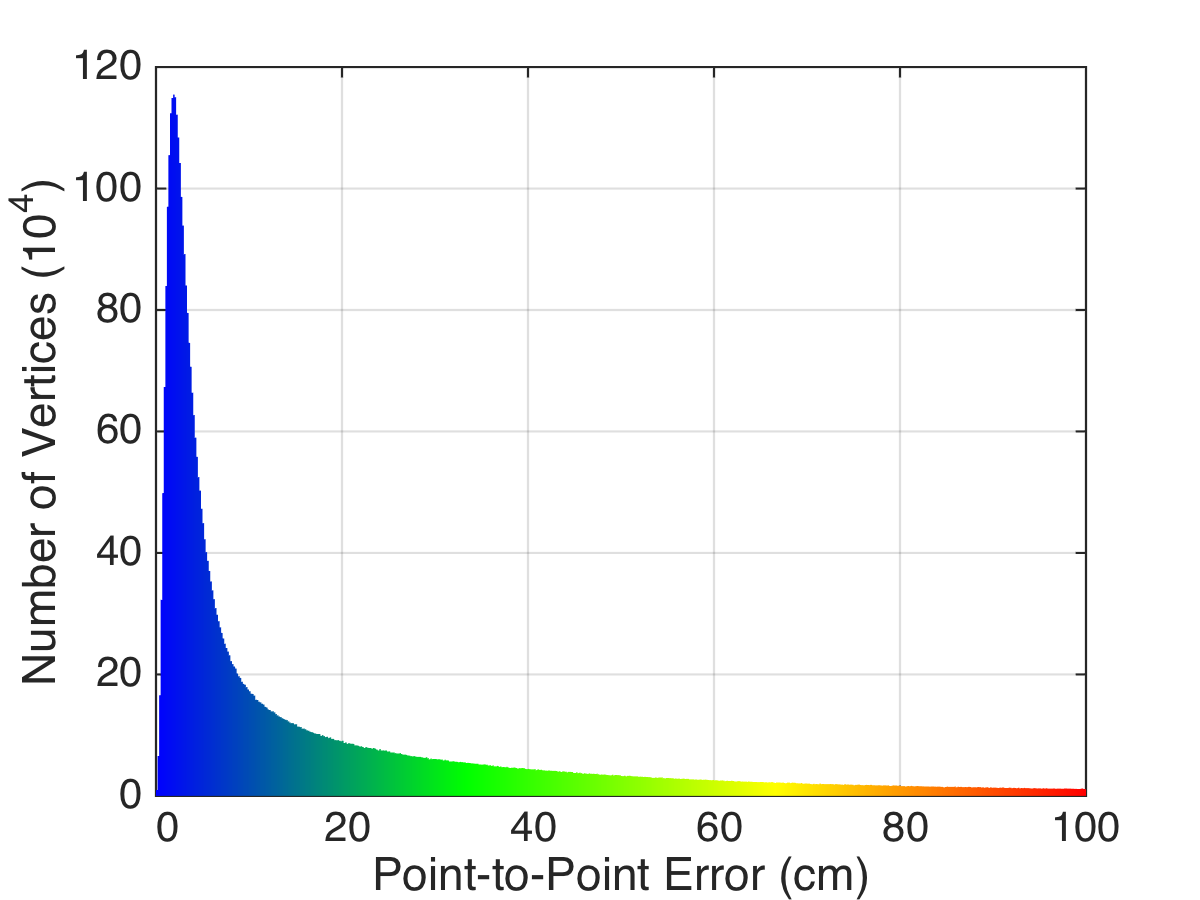}
      \put(30,60){\footnotesize median $=$ 8.43~cm}
      \put(30,50){\footnotesize 75\% $=$ 32.22~cm}
    \end{overpic}
    }
    \subfigure[Sequence 09 (With regularization)]{
    \begin{overpic}[width=0.5\columnwidth]{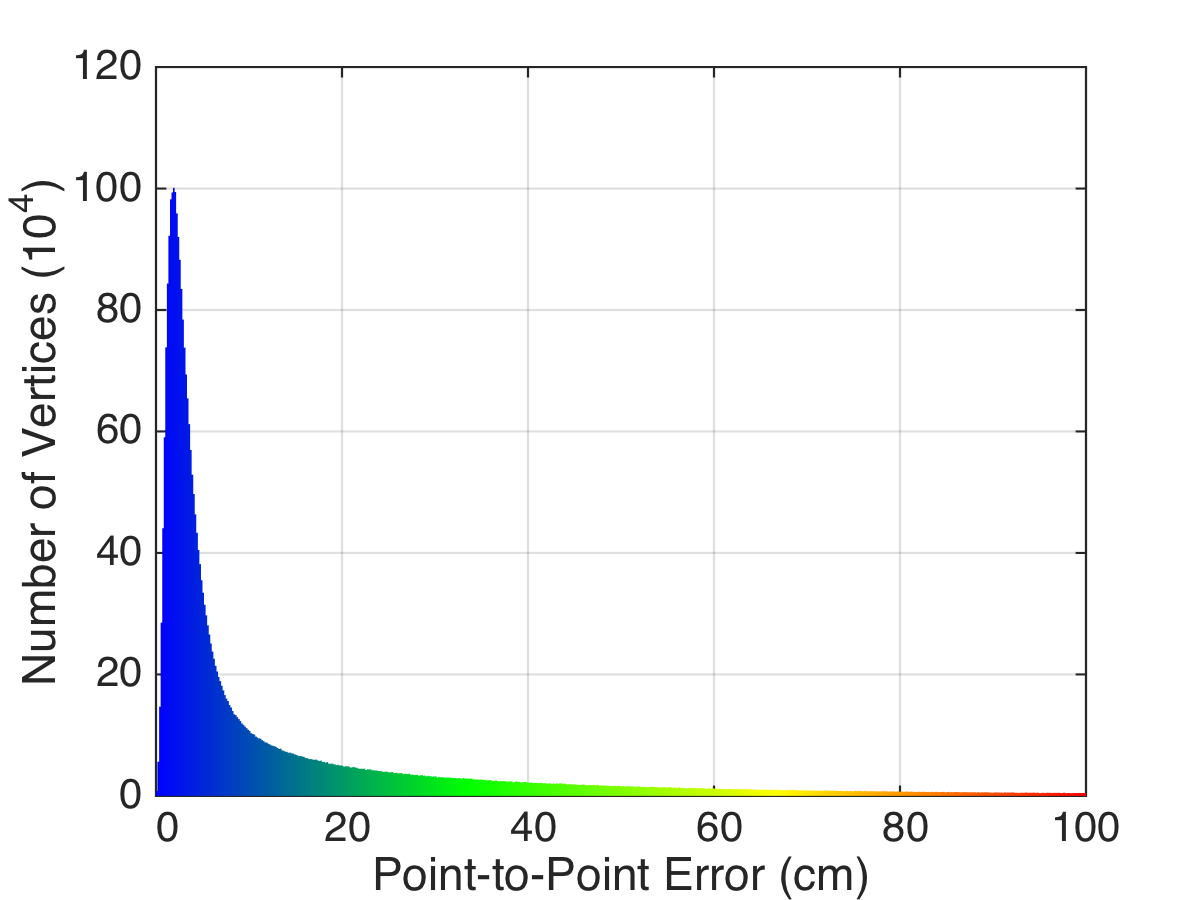}
      \put(30,60){\footnotesize median $=$ 5.04~cm}
      \put(30,50){\footnotesize 75\% $=$ 19.06~cm}
    \end{overpic}
    \includegraphics[width=0.45\columnwidth]{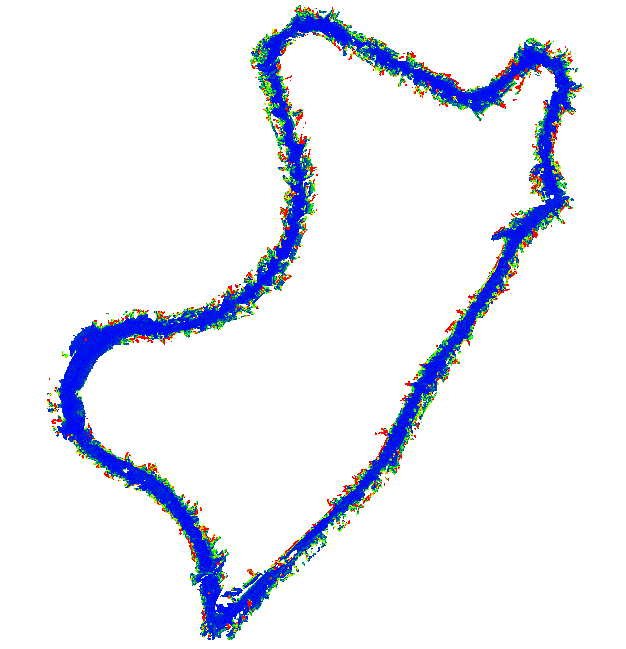}
    }
  }
  \caption{
  A summary of the dense reconstruction quality for three scenarios (one scenario per row) from the KITTI-VO public benchmark dataset.
  The left side are the results before regularization and the right side are after regularization.
  Next to each histogram of point-to-point errors is a top-view, colored reconstruction errors corresponding to the same colors in the histogram.
  The regularizer reduces the reconstruction's error approximately 40\%, primarily by removing uncertain surfaces --- as can be seen when you contrast the raw (far left) and regularized (far right) reconstruction errors.}
  \label{fig:detailed_error_metrics}
\end{figure*}

\begin{figure*}[t!]
  \centering{
    \subfigure[Sequence 00: The maps accurately depict complex urban environments, including a variety of trees and automobiles.  However, gaps do exist in the reconstruction when an area is not directly observed (e.g., behind a vehicle or on the inside corner of a turn). The left image is the full 3.7~km trajectory.]{
    \includegraphics[width=0.5\columnwidth]{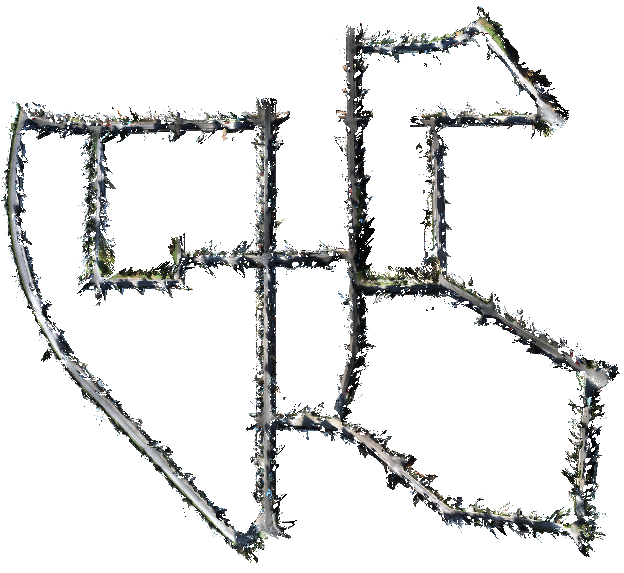}
    \includegraphics[width=0.5\columnwidth]{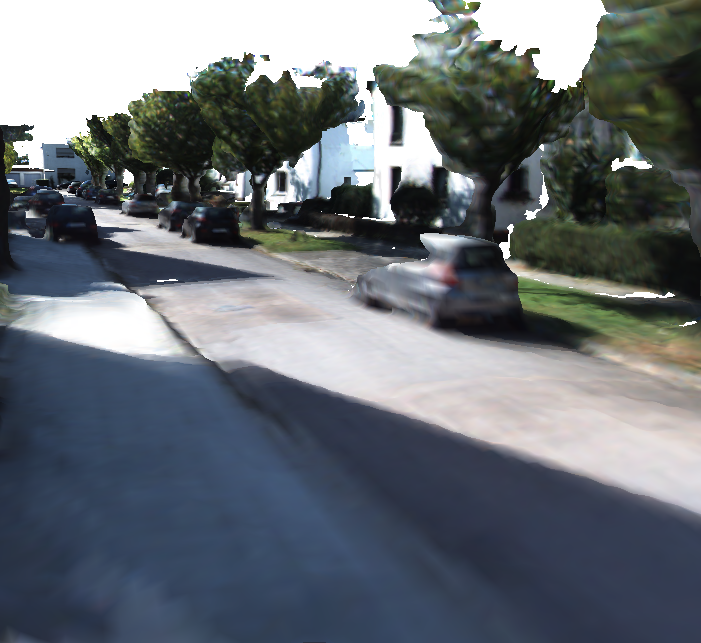}
    \includegraphics[width=0.5\columnwidth]{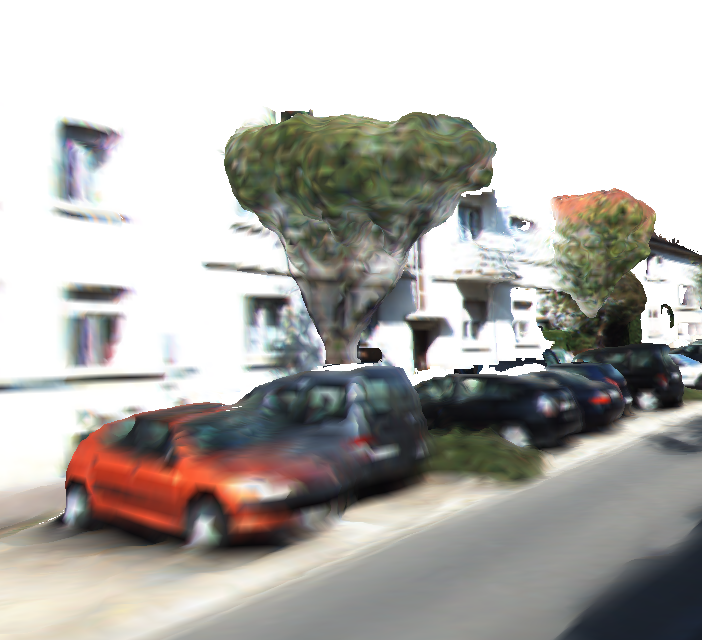}
    \includegraphics[width=0.5\columnwidth]{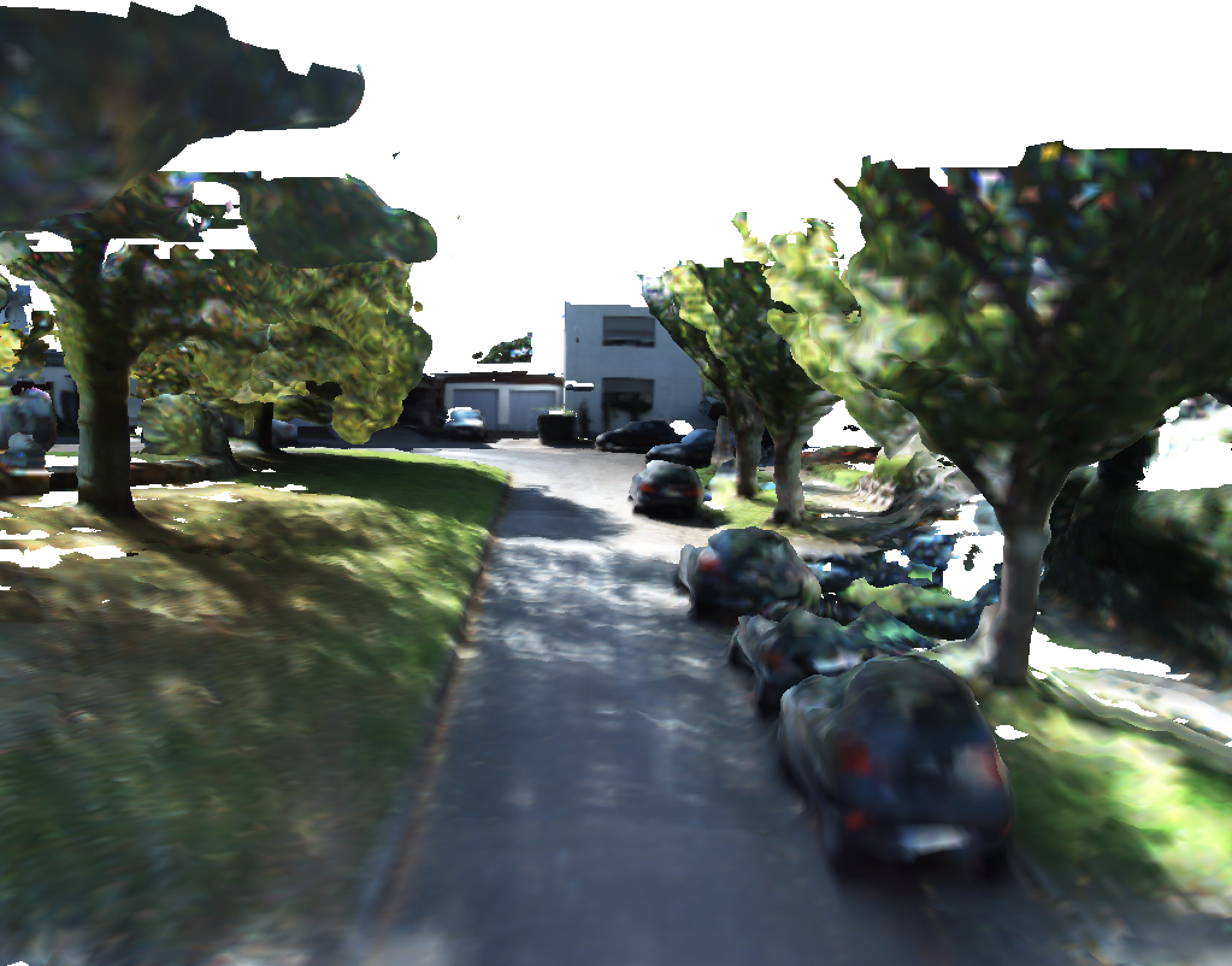}
    }
    
    \subfigure[Sequence 07: Buildings and vegetation are accurately modelled in our system.]{
    \includegraphics[width=0.5\columnwidth]{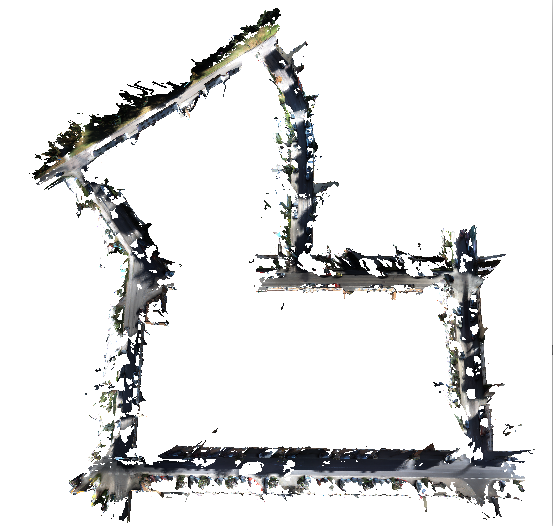}
    \includegraphics[width=0.5\columnwidth]{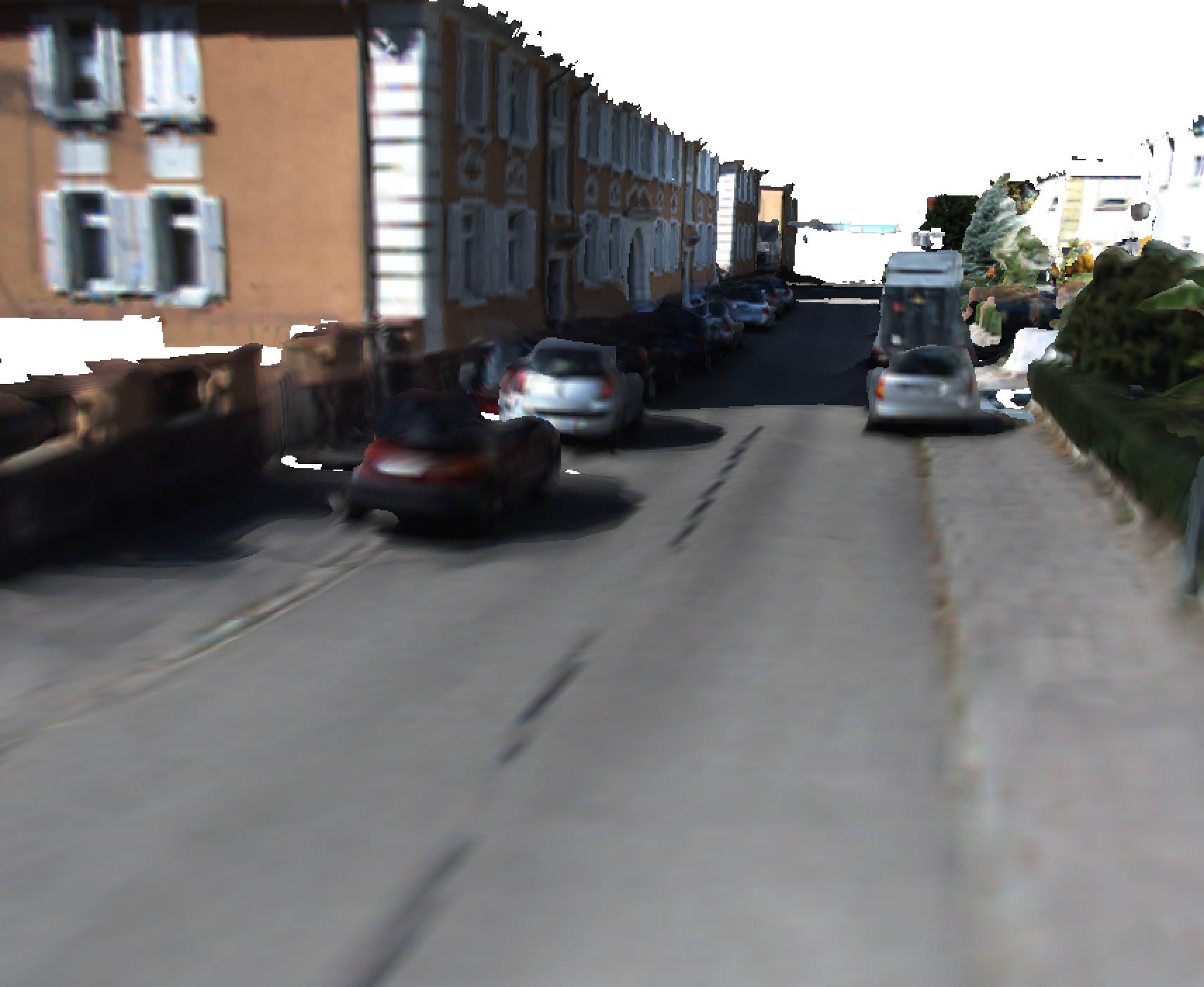}
    \includegraphics[width=0.5\columnwidth]{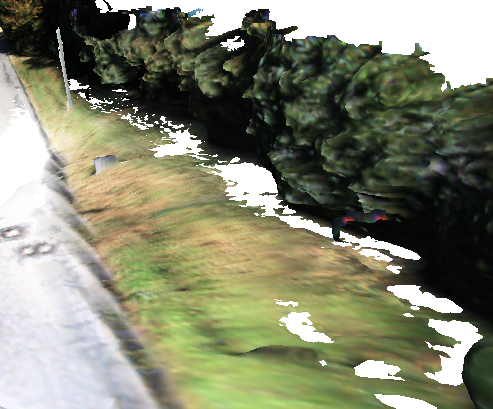}
    \includegraphics[width=0.5\columnwidth]{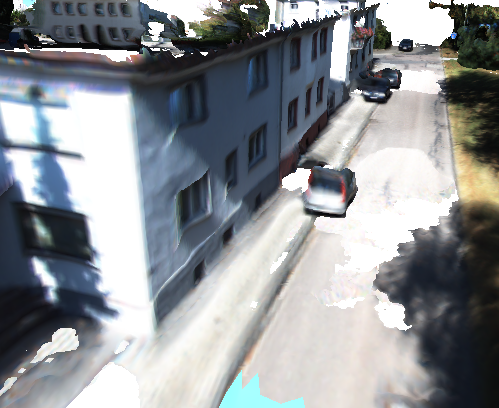}
    }
    
    \subfigure[Sequence 09: The forward-facing camera sometimes sees dynamic objects (e.g., car in the center of the right image) which are only observed in one image frame and therefore may clutter the dense reconstruction. However, the overall reconstruction is still quite accurate, even in the presence of vegetation.]{
    \includegraphics[width=0.5\columnwidth]{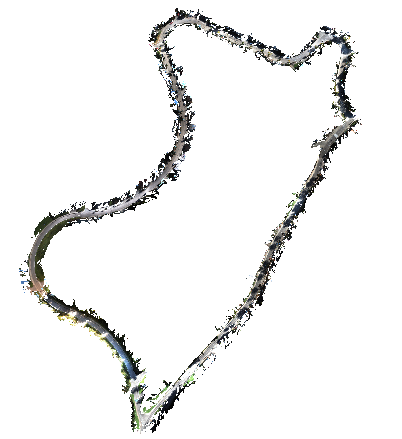}
    \includegraphics[width=0.5\columnwidth]{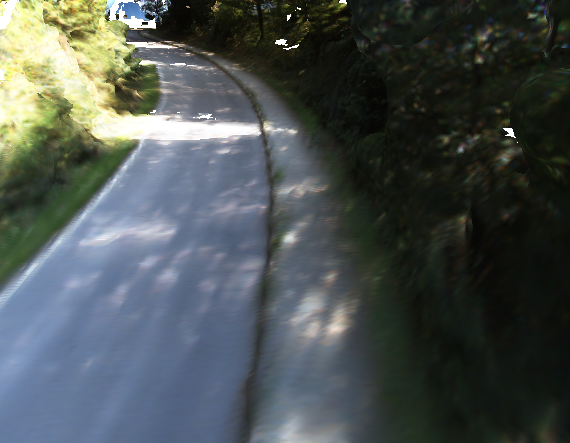}
    \includegraphics[width=0.5\columnwidth]{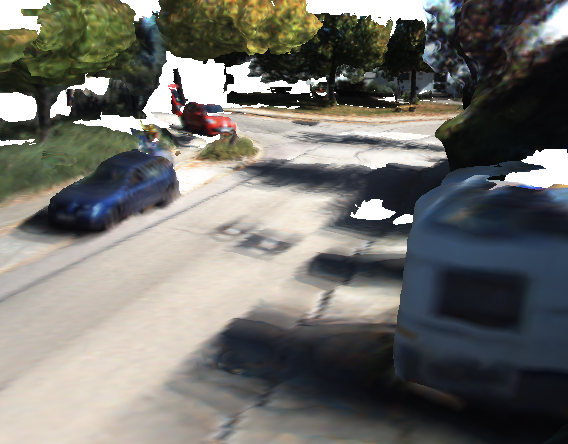}
    \includegraphics[width=0.5\columnwidth]{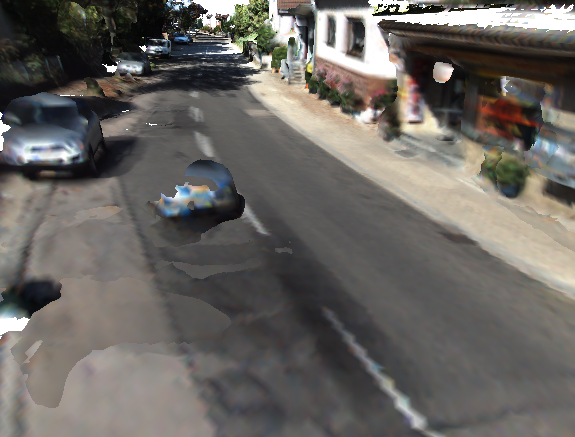}
    }
  }
  \caption{
  A few representative sample images for various points of views (offset from the original camera's position) along each trajectory.
  All sample images are of the final regularized reconstruction with 10~cm voxels.
  The submitted video provides full fly-through footage for each sequence.
  }
  \label{fig:sample_images}
\end{figure*}

\section{Depth-Map Estimation}\label{sec:depth_maps}

\ignore{
In the results section, we evaluate our DENSER system in a particularly challenging setting: obtaining rich and dense models of a town from a single forward-facing stereo camera mounted on top of a car is minimal for two reasons.
First, visual odometry (VO) algorithms accurately compute the egomotion of a vehicle from a stream of images \cite{cvisic2015stereo}\cite{mur2015orb}\cite{engel2015large}.
In particular, in our approach, we use a stereo camera to recover the scale of the map.
Second, unlike traditional sparse 3D reconstructions that detect a few key-points per image, a recent trend of new dense mapping algorithms has emerged that enables the creation of increasingly accurate dense disparity maps \cite{hermann2013iterative}\cite{piniesFSR2015}\cite{yamaguchi2012continuous} and therefore richer models of the environment. 
}

This section describes our module that produces a stream of depth maps ($D$) which are the input to the previously-explained fusion scheme.
The algorithm implemented, based on \cite{ranftl2012pushing}, minimizes an energy functional that contains, as in Equation \ref{eq:vm_energy_tv_simple}, a regularization term which assumes a prior model about the structure of the scene and a data term that measures the similarity between corresponding pixels in the stereo images.

\ignore{
This algorithm was chosen after testing similar variational approaches because it offers a good trade-off between speed and high quality depth maps.
An important lesson we learnt during this process was focus on four key aspects when generating depth maps:
(1) choose the fidelity/similarity metric between pixels for the data term,
(2) choose the regularization functional containing prior knowledge about the scene,
(3) use the appearance cues to identify objects, and
(4) use optimization tricks to avoid bad local minima.
}

\subsection{Census Transform Signature Data Term}

The data term is given by:

\begin{equation}
E_{data}(d; I_L, I_R) = \iint_{\Omega} |\rho(d, x, y)| dx\,dy
\end{equation}

\noindent where the coordinates are $(x,y)$ for a particular pixel in the reference image, and the function $\rho(d, x, y) = Sim^W(I_L(x+d,y), I_R(x,y))$ measures the similarity between two pixels using a window of size $W$ for a candidate disparity $d \in D$.
In this work we use a non-parametric local transform known as the Census Transform Signature (CTS) \cite{ZabihW94} as our similarity metric. This metric has been shown to be both illumination invariant and fast to compute.
Given a pixel, the CTS computes a bit string by comparing the chosen pixel with the local window $W$ centered around it.
A bit is set to $1$ if the corresponding pixel has a lower intensity than the pixel of interest.
The similarity measure between two windows is then given by the Hamming distance between the two bit strings defining them.

\subsection{Affine Regularization}

For ill-posed problems, like depth-map estimation, good and apt priors are essential --- whether the prior is task-specific and bespoke \cite{guney2015displets} or more general.
A common choice is to use TV regularization as a prior to favor piecewise constant solutions.
However, its use lends to poor depth-map estimates over outdoor sequences by creating fronto-parallel surfaces.
Figure~\ref{fig:tv_tgv_tensor} shows some of the artifacts created after back-projecting the point cloud for planar surfaces not orthogonal to the image plane (e.g. the roads and walls which dominate our urban scenes).
Thus we reach for a Total Global Variation (TGV) regularization term which favours planar surfaces in any orientation:

\begin{equation}\label{eq:reg_2D}
E_{reg}(d) = \min_{\mathbf{w} \in \mathbb{R}^2} \alpha_1 \iint_{\Omega} |\mathbf{T}\  \nabla d - \mathbf{w}| dx\:dy  + \alpha_2 \iint_{\Omega} |\nabla \mathbf{w}| dx\:dy
\end{equation}

\noindent where $\mathbf{w}$ allows the disparity $d$ in a region of the depth map to change at a constant rate and therefore creates planar surfaces with different orientations.
The meaning of matrix $\mathbf{T}$ will be explained in the next subsection.

\subsection{Leveraging Appearance}

A common problem that arises during the energy minimization is the tension between preserving object discontinuities while respecting the smoothness prior.
Ideally the solutions show preserve intra-object continuity and inter-object discontinuity. 

One may mitigate this tension by using the isotropic appearance gradient ($\nabla I$) as an indicator of boundaries between objects by defining  $\mathbf{T}$ from Equation~\ref{eq:reg_2D}, as:

\begin{equation}
\mathbf{T'} = \exp(-\gamma|\nabla I|^{\beta})
\end{equation}

However, though this aids the regularizer, it does not contain information about the direction of the border between the objects.
To take this information into account  we adopt an anisotropic (as opposed to isotropic) diffusion tensor given by:  

\begin{equation}
\mathbf{T} = \exp(-\gamma|\nabla I|^{\beta}) n n^T + n^{\perp} n^{\perp^T}
\end{equation}

\noindent where $n = \frac{\nabla I}{|\nabla I|}$ and $n^\perp$ is its orthogonal complement.
The effect of this tensor on Equation~\ref{eq:reg_2D} is to decompose the components of the disparity's gradient ($\nabla d$) in the directions aligned with $n$ and $n^\perp$.
We do not penalize large gradient components aligned with $n$ using the exponential; we do penalize components aligned with $n^\perp$.
In other words, if there is a discontinuity visible in the color image, then it is highly probable that there is a discontinuity in the depth image.
The benefits of this tensor term are visually depicted in Figure~\ref{fig:tv_tgv_tensor}.

\section{Results}\label{sec:results}


This section provides an extensive analysis of our system, the parameters for which are provided in Table~\ref{tab:parameters}.
We use the publicly available KITTI dataset \cite{Geiger:2012:Areweready}.
For ground truth we consolidated all Velodyne HDL-64E laser scans into a single reference frame.
We  keep in mind though  that this is not a perfect ground truth because of inevitable errors in the data-set pose estimates (we observed up to 3~m of vertical drift throughout Sequence 09). 

Three sequences were selected: 00 which  is a 3.7~km route, but only the first 1.0~km is quantitatively evaluated to avoid poor loop closures in KITTI's GPS/INS-based ground-truth poses, 07, and 09.
A summary of the physical scale of each is provided in Table~\ref{tab:kitti_vo_summary}, along with the total time required to fuse data into our HVG structure with either 10~cm or 20~cm voxels using an GeForce GTX TITAN with 6~GiB.

We first processed all scenarios with 10~cm voxels and compared the dense reconstruction model, both before and after regularization, to the laser scans, see Table~\ref{tab:kitti_vo_errors_10cm}.
The regularizer on average reduced the median error by 40\% (10~cm $\rightarrow$ 6~cm), the 75-percentile error by 36\% (36~cm $\rightarrow$ 23~cm), and the surface area by 32\% (55,008~m$^2$ $\rightarrow$ 37,336~m$^2$).
\ignore{
\pmnnotes{we need to be much more compelling here...suggestions to write about}

\begin{itemize}
    \item how long does it take..
    \item what percentage saving is achieved by the compressed data structure (over naive)
    \item where do we still fail and why - thats important. Right now we are only positive. Can we show a challenging case ?
\end{itemize}
}
In these large-scale reconstructions, the compressed voxel grid structure provides near real-time fusion performance (Table~\ref{tab:kitti_vo_summary}) while vastly increasing the size of reconstructions.
The legacy voxel grid was only able to process 205~m; this stands in stark contrast to the 1.6~km reconstructed with the HVG for the same amount GPU memory.

This is further detailed in Figure~\ref{fig:detailed_error_metrics} where it becomes clear that errors in the initial ``raw'' fusion largely come from false surfaces created in the input depth map caused by sharp discontinuities in the original image.
The regularizer removes many of these surfaces, which dominate the tail of the histogram plots and are visible as red points in the point-cloud plots.
When processed at 20~cm voxel resolution, the results are similar, though with slightly higher error metrics (as would be expected) and are shown in Table~\ref{tab:kitti_vo_errors_20cm}.

Figure~\ref{fig:sample_images}, shows the bird's-eye view of each sequence with representative snapshots of the reconstruction showing both where we performed well and where the system struggled.
To illustrate the quality of the reconstructions, we selected several snapshots from camera viewpoints offset in both translation and rotation to the original stereo camera position, thereby providing an accurate depiction of the 3D structure.
Overall, the reconstructions are quite visually appealing; however, some artifacts such as holes are persistent in regions with poor texture or with large changes in illumination. This is an expected result since, in these cases, no depth map can be accurately inferred.
The submitted video provides a fly-through of each sequence to visualize the quality of our final regularized 3D reconstructions.

\section{Conclusion}\label{sec:conclusion}

We presented a state-of-the-art dense mapping system for city-scale dense reconstructions.
We overcame the primary technical challenge of regularising voxel data in the HVG-compressed 3D structure by redefining the gradient and divergence operators to account for the additional boundary conditions introduced by the data structure.
This both \emph{enables} regularization and prevents the regularizer from erroneously \emph{extrapolating} surface data.
We evaluated our system's accuracy, for different granularities, against  3.4~km of laser data data.
Our regularizer consistently reduced the reconstruction error metrics by 40\%, for a median accuracy of 6~cm over $2.8e5$~m$^2$ of constructed area.
Though computational requirements of the regularization steps impose run-time constraints, our dense fusion system runs in real-time when laser data is our range input.



\bibliographystyle{IEEEtran}
\bibliography{references}

\begin{thebibliography}{10}
\providecommand{\url}[1]{#1}
\csname url@samestyle\endcsname
\providecommand{\newblock}{\relax}
\providecommand{\bibinfo}[2]{#2}
\providecommand{\BIBentrySTDinterwordspacing}{\spaceskip=0pt\relax}
\providecommand{\BIBentryALTinterwordstretchfactor}{4}
\providecommand{\BIBentryALTinterwordspacing}{\spaceskip=\fontdimen2\font plus
\BIBentryALTinterwordstretchfactor\fontdimen3\font minus
  \fontdimen4\font\relax}
\providecommand{\BIBforeignlanguage}[2]{{%
\expandafter\ifx\csname l@#1\endcsname\relax
\typeout{** WARNING: IEEEtran.bst: No hyphenation pattern has been}%
\typeout{** loaded for the language `#1'. Using the pattern for}%
\typeout{** the default language instead.}%
\else
\language=\csname l@#1\endcsname
\fi
#2}}
\providecommand{\BIBdecl}{\relax}
\BIBdecl

\bibitem{pollefeys2008}
M.~Pollefeys, D.~Nist{\'e}r, J.-M. Frahm, A.~Akbarzadeh, P.~Mordohai, B.~Clipp,
  C.~Engels, D.~Gallup, S.-J. Kim, P.~Merrell, C.~Salmi, S.~Sinha, B.~Talton,
  L.~Wang, Q.~Yang, H.~Stew{\'e}nius, R.~Yang, G.~Welch, and H.~Towles,
  ``Detailed real-time urban 3d reconstruction from video,''
  \emph{International Journal of Computer Vision}, vol.~78, no. 2-3, pp.
  143--167, July 2008.

\bibitem{FurukawaICCV2009}
S.~Agarwal, N.~Snavely, I.~Simon, S.~M. Seitz, and R.~Szeliski, ``Building rome
  in a day,'' in \emph{Twelfth IEEE International Conference on Computer Vision
  (ICCV 2009)}.\hskip 1em plus 0.5em minus 0.4em\relax Kyoto, Japan: IEEE,
  September 2009.

\bibitem{FurukawaCSS10}
Y.~Furukawa, B.~Curless, S.~M. Seitz, and R.~Szeliski, ``Towards internet-scale
  multi-view stereo,'' in \emph{The Twenty-Third {IEEE} Conference on Computer
  Vision and Pattern Recognition, {CVPR} 2010, San Francisco, CA, USA, 13-18
  June 2010}, 2010, pp. 1434--1441.

\bibitem{Chambolle:2011:FirstOrderPrimalDualAlgorithm}
A.~{Chambolle} and T.~{Pock}, ``{A} {First}-{Order} {Primal}-{Dual} {Algorithm}
  for {Convex} {Problems} with {Applications} to {Imaging},'' \emph{{J}.
  {Math}. {Imaging} {Vis}.}, vol.~40, no.~1, pp. 120--145, May 2011.

\bibitem{Goldluecke:2012:naturalvectorialtotal}
B.~{Goldluecke}, E.~{Strekalovskiy}, and D.~{Cremers}, ``The natural vectorial
  total variation which arises from geometric measure theory,'' \emph{{SIAM}
  {Journal} on {Imaging} {Sciences}}, vol.~5, no.~2, pp. 537--563, 2012.

\bibitem{Klingensmith_2015_7924}
M.~Klingensmith, I.~Dryanovski, S.~Srinivasa, and J.~Xiao, ``Chisel: Real time
  large scale 3d reconstruction onboard a mobile device,'' in \emph{Robotics
  Science and Systems 2015}, July 2015.

\bibitem{engel2015_stereo_lsdslam}
J.~Engel, J.~Stueckler, and D.~Cremers, ``Large-scale direct slam with stereo
  cameras,'' in \emph{International Conference on Intelligent Robots and
  Systems (IROS)}, 2015.

\bibitem{vineet:etal:icra2015}
V.~{Vineet}, O.~{Miksik}, M.~{Lidegaard}, M.~{Niessner}, S.~{Golodetz},
  V.~{Prisacariu}, O.~{K{\"a}hler}, D.~{Murray}, S.~{Izadi}, P.~{Peerez}, and
  P.~{Torr}, ``Incremental dense semantic stereo fusion for large-scale
  semantic scene reconstruction,'' in \emph{{IEEE} {International} {Conference}
  on {Robotics} and {Automation} ({ICRA} 2015)}, May 2015, pp. 75--82.

\bibitem{Newcombe:2011:KinectFusionRealtimedense}
R.~A. {Newcombe}, A.~J. {Davison}, S.~{Izadi}, P.~{Kohli}, O.~{Hilliges},
  J.~{Shotton}, D.~{Molyneaux}, S.~{Hodges}, D.~{Kim}, and A.~{Fitzgibbon},
  ``{KinectFusion}: {Real}-time dense surface mapping and tracking,'' in
  \emph{Mixed and augmented reality ({ISMAR}), 2011 10th {IEEE} international
  symposium on}.\hskip 1em plus 0.5em minus 0.4em\relax {IEEE}, 2011, pp.
  127--136.

\bibitem{schoeps20153dv}
T.~Sch\"ops, T.~Sattler, C.~H\"ane, and M.~Pollefeys, ``{3D} modeling on the
  go: Interactive {3D} reconstruction of large-scale scenes on mobile
  devices,'' in \emph{International Conference on 3D Vision (3DV)}, 2015.

\bibitem{Whelan:2014:Realtimelargescaledense}
T.~{Whelan}, M.~{Kaess}, H.~{Johannsson}, M.~{Fallon}, J.~J. {Leonard}, and
  J.~{McDonald}, ``Real-time large-scale dense {RGB}-{D} {SLAM} with volumetric
  fusion,'' \emph{The {International} {Journal} of {Robotics} {Research}}, p.
  0278364914551008, Dec. 2014.

\bibitem{Curless:1996:volumetricmethodbuilding}
B.~{Curless} and M.~{Levoy}, ``{A} volumetric method for building complex
  models from range images,'' in \emph{Proceedings of the 23rd annual
  conference on {Computer} graphics and interactive techniques}.\hskip 1em plus
  0.5em minus 0.4em\relax {ACM}, 1996, pp. 303--312, 02090.

\bibitem{Pradeep:2013:MonoFusionRealtime3D}
V.~{Pradeep}, C.~{Rhemann}, S.~{Izadi}, C.~{Zach}, M.~{Bleyer}, and
  S.~{Bathiche}, ``{MonoFusion}: {Real}-time {3D} reconstruction of small
  scenes with a single web camera,'' in \emph{2013 {IEEE} {International}
  {Symposium} on {Mixed} and {Augmented} {Reality} ({ISMAR})}, Oct. 2013, pp.
  83--88.

\bibitem{Alcantarilla14ppniv}
P.~F. Alcantarilla, C.~Beall, and F.~Dellaert, ``Large-scale dense 3d
  reconstruction from stereo imagery,'' in \emph{In 5th Workshop on Planning,
  Perception and Navigation for Intelligent Vehicles (PPNIV)}, Tokyo, Japan,
  11/2013 2013.

\bibitem{mur2015orb}
R.~Mur-Artal, J.~Montiel, and J.~D. Tardos, ``Orb-slam: a versatile and
  accurate monocular slam system,'' \emph{arXiv preprint arXiv:1502.00956},
  2015.

\bibitem{RSSYguel07}
M.~Yguel, C.~T.~M. Keat, C.~Braillon, C.~Laugier, and O.~Aycard, ``Dense
  mapping for range sensors: Efficient algorithms and sparse representations,''
  in \emph{Robotics: Science and Systems III, June 27-30, 2007, Georgia
  Institute of Technology, Atlanta, Georgia, {USA}}, 2007.

\bibitem{Zeng:2013:Octreebasedfusionrealtime}
M.~{Zeng}, F.~{Zhao}, J.~{Zheng}, and X.~{Liu}, ``Octree-based fusion for
  realtime {3D} reconstruction,'' \emph{Graphical {Models}}, vol.~75, no.~3,
  pp. 126--136, 2013, 00029.

\bibitem{Keller:2013:Realtime3dreconstruction}
M.~{Keller}, D.~{Lefloch}, M.~{Lambers}, S.~{Izadi}, T.~{Weyrich}, and
  A.~{Kolb}, ``Real-time 3d reconstruction in dynamic scenes using point-based
  fusion,'' in \emph{{3DTV}-{Conference}, 2013 {International} {Conference}
  on}.\hskip 1em plus 0.5em minus 0.4em\relax {IEEE}, 2013, pp. 1--8.

\bibitem{Teschner:2003:OptimizedSpatialHashing}
M.~{Teschner}, B.~{Heidelberger}, M.~{Mueller}, D.~{Pomeranets}, and
  M.~{Gross}, ``Optimized {Spatial} {Hashing} for {Collision} {Detection} of
  {Deformable} {Objects},'' 2003, pp. 47--54.

\bibitem{Kim:2009:Multiviewimagetof}
Y.~M. {Kim}, C.~{Theobalt}, J.~{Diebel}, J.~{Kosecka}, B.~{Miscusik}, and
  S.~{Thrun}, ``Multi-view image and tof sensor fusion for dense 3d
  reconstruction,'' in \emph{Computer {Vision} {Workshops} ({ICCV}
  {Workshops}), 2009 {IEEE} 12th {International} {Conference} on}.\hskip 1em
  plus 0.5em minus 0.4em\relax {IEEE}, 2009, pp. 1542--1549.

\bibitem{Newcombe:2011:DTAMDensetracking}
R.~A. {Newcombe}, S.~J. {Lovegrove}, and A.~J. {Davison}, ``{DTAM}: {Dense}
  tracking and mapping in real-time,'' in \emph{Computer {Vision} ({ICCV}),
  2011 {IEEE} {International} {Conference} on}.\hskip 1em plus 0.5em minus
  0.4em\relax {IEEE}, 2011, pp. 2320--2327, 00384.

\bibitem{Geiger2010ACCV}
A.~Geiger, M.~Roser, and R.~Urtasun, ``Efficient large-scale stereo matching,''
  in \emph{Asian Conference on Computer Vision (ACCV)}, 2010.

\bibitem{ranftl2012pushing}
R.~Ranftl, S.~Gehrig, T.~Pock, and H.~Bischof, ``Pushing the limits of stereo
  using variational stereo estimation,'' in \emph{Intelligent Vehicles
  Symposium (IV), 2012 IEEE}.\hskip 1em plus 0.5em minus 0.4em\relax IEEE,
  2012, pp. 401--407.

\bibitem{ferstl2014b}
D.~Ferstl, C.~Reinbacher, G.~Riegler, M.~Ruether, and H.~Bischof, ``atgv-sf:
  Dense variational scene flow through projective warping and higher order
  regularization,'' in \emph{Proceedings of International Conference on 3D
  Vision (3DV)}, December 2014.

\bibitem{ZabihW94}
R.~Zabih and J.~Woodfill, ``Non-parametric local transforms for computing
  visual correspondence,'' in \emph{Computer Vision - ECCV'94, Third European
  Conference on Computer Vision, Stockholm, Sweden, May 2-6, 1994, Proceedings,
  Volume {II}}, 1994, pp. 151--158.

\bibitem{Niessner:2013:Realtime3DReconstruction}
M.~{Nie{\ss}ner}, M.~{Zollh{\"o}fer}, S.~{Izadi}, and M.~{Stamminger},
  ``Real-time {{3D Reconstruction}} at {{Scale Using Voxel Hashing}},''
  \emph{ACM Trans. Graph.}, vol.~32, no.~6, pp. 169:1--169:11, Nov. 2013,
  00050.

\bibitem{Tanner:2015:BOR2GBuildingOptimal}
M.~{Tanner}, P.~{Pini{\'e}s}, L.~M. {Paz}, and P.~{Newman}, ``{BOR2G}:
  {Building} {Optimal} {Regularised} {Reconstructions} with {GPUs} (in
  cubes),'' in \emph{International {Conference} on {Field} and {Service}
  {Robotics} ({FSR})}, Toronto, {ON}, {Canada}, Jun. 2015.

\bibitem{Whelan:2012:KintinuousSpatiallyExtended}
T.~{Whelan}, M.~{Kaess}, M.~F. {Fallon}, H.~{Johannsson}, J.~J. {Leonard}, and
  J.~B. {McDonald}, ``Kintinuous: {{Spatially Extended KinectFusion}},'' in
  \emph{{{RSS Workshop}} on {{RGB-D}}: {{Advanced Reasoning}} with {{Depth
  Cameras}}}, Sydney, Australia, Jul. 2012.

\bibitem{Rockafellar:1970:ConvexAnalysis}
R.~T. {Rockafellar}, \emph{Convex {{Analysis}}}.\hskip 1em plus 0.5em minus
  0.4em\relax Princeton, New Jersey: {Princeton University Press}, 1970.

\bibitem{guney2015displets}
F.~G{\"u}ney and A.~Geiger, ``Displets: Resolving stereo ambiguities using
  object knowledge,'' in \emph{Proceedings of the IEEE Conference on Computer
  Vision and Pattern Recognition}, 2015, pp. 4165--4175.

\bibitem{Geiger:2012:Areweready}
A.~{Geiger}, P.~{Lenz}, and R.~{Urtasun}, ``Are we ready for {{Autonomous
  Driving}}? {{The KITTI Vision Benchmark Suite}},'' in \emph{Conference on
  {{Computer Vision}} and {{Pattern Recognition}} ({{CVPR}})}, 2012, 00627.

\end{thebibliography}
~ 

\end{document}